%% file: calibration-arxiv.tex
\documentclass[11pt]{article}
\usepackage[margin=1in]{geometry}
\usepackage[round, compress]{natbib}

\usepackage{microtype}
\usepackage{graphicx}
\usepackage{subcaption}
\usepackage{booktabs} 
\usepackage{tcolorbox}
\usepackage{stackrel}

\usepackage{hyperref}

\usepackage{algorithm}
\usepackage{algorithmic}

\usepackage{yub}
\usepackage{smile}
\usepackage{empheq}

\allowdisplaybreaks

\usepackage[normalem]{ulem}
\usepackage{xcolor}

\def\logistic{{\rm logistic}}
\def\uc{{\rm underconf}}
\def\CE{{\rm CE}}

\hypersetup{
    colorlinks,
    linkcolor={blue!50!black},
    citecolor={blue!50!black},
}
\colorlet{linkequation}{blue}

\def\shownotes{0}  
\ifnum\shownotes=1
\newcommand{\authnote}[2]{{\scriptsize $\ll$\textsf{#1 notes: #2}$\gg$}}
\else
\newcommand{\authnote}[2]{}
\fi

\def\normbar{R_\star}
\def\cosbar{c_\star}

\def\deltapcal{{\Delta_p^{\sf cal}}}

\def\barsigma{{\bar \sigma}}
\def\baralpha{{\bar \alpha}}
\def\barlambda{{\bar \lambda}}
\def\bF{{\boldsymbol F}}
\def\bp{{\boldsymbol p}}

\def\de{{\rm d}}
\def\bJ{{\boldsymbol J}}

\title{Don’t Just Blame Over-parametrization for Over-confidence:
  \\ Theoretical Analysis of Calibration in Binary Classification
}

\author{
  Yu Bai\thanks{Salesforce Research. E-mail:~\texttt{yu.bai@salesforce.com}}
  \and
  Song Mei\thanks{University of California, Berkeley. E-mail:~\texttt{songmei@berkeley.edu}}
  \and
  Huan Wang\thanks{Salesforce Research. E-mail:~\texttt{\{huan.wang, cxiong\}@salesforce.com}}
  \and
  Caiming Xiong\footnotemark[3]
}
\date{\today}

\begin{document}

\maketitle

\def\sec{Sections-arxiv}

\input{\sec/abstract.tex}
\input{\sec/intro.tex}

\input{\sec/prelim.tex}
\input{\sec/logistic.tex}
\input{\sec/general.tex}
\input{\sec/experiments.tex}

\input{\sec/proof-sketch.tex}
\input{\sec/conclusion.tex}
\input{\sec/acknowledgment.tex}

\bibliography{bib}
\bibliographystyle{plainnat}

\makeatletter
\def\renewtheorem#1{%
  \expandafter\let\csname#1\endcsname\relax
  \expandafter\let\csname c@#1\endcsname\relax
  \gdef\renewtheorem@envname{#1}
  \renewtheorem@secpar
}
\def\renewtheorem@secpar{\@ifnextchar[{\renewtheorem@numberedlike}{\renewtheorem@nonumberedlike}}
\def\renewtheorem@numberedlike[#1]#2{\newtheorem{\renewtheorem@envname}[#1]{#2}}
\def\renewtheorem@nonumberedlike#1{  
\def\renewtheorem@caption{#1}
\edef\renewtheorem@nowithin{\noexpand\newtheorem{\renewtheorem@envname}{\renewtheorem@caption}}
\renewtheorem@thirdpar
}
\def\renewtheorem@thirdpar{\@ifnextchar[{\renewtheorem@within}{\renewtheorem@nowithin}}
\def\renewtheorem@within[#1]{\renewtheorem@nowithin[#1]}
\makeatother

\renewtheorem{theorem}{Theorem}[section]
\renewtheorem{lemma}{Lemma}[section]
\renewtheorem{remark}{Remark}
\renewtheorem{corollary}{Corollary}[section]
\renewtheorem{observation}{Observation}[section]
\renewtheorem{proposition}{Proposition}[section]
\renewtheorem{definition}{Definition}[section]
\renewtheorem{claim}{Claim}[section]
\renewtheorem{fact}{Fact}[section]
\renewtheorem{assumption}{Assumption}[section]
\renewcommand{\theassumption}{\Alph{assumption}}
\renewtheorem{conjecture}{Conjecture}[section]

\appendix

\input{\sec/tools.tex}
\input{\sec/closed-form.tex}
\input{\sec/proof-proportional.tex}
\input{\sec/proofs.tex}
\input{\sec/appendix-experiments.tex}

\end{document}

%% file: Sections-arxiv/abstract.tex
\begin{abstract}
  Modern machine learning models with high accuracy are often miscalibrated---the predicted top probability does not reflect the actual accuracy, and tends to be \emph{over-confident}. It is commonly believed that such over-confidence is mainly due to \emph{over-parametrization}, in particular when the model is large enough to memorize the training data and maximize the confidence.

  In this paper, we show theoretically that over-parametrization is not the only reason for over-confidence. We prove that \emph{logistic regression is inherently over-confident}, in the realizable, under-parametrized setting where the data is generated from the logistic model, and the sample size is much larger than the number of parameters. Further, this over-confidence happens for general well-specified binary classification problems as long as the activation is symmetric and concave on the positive part. Perhaps surprisingly, we also show that over-confidence is not always the case---there exists another activation function (and a suitable loss function) under which the learned classifier is \emph{under-confident} at some probability values. Overall, our theory provides a precise characterization of calibration in realizable binary classification, which we verify on simulations and real data experiments.
\end{abstract}




%% file: Sections-arxiv/intro.tex
\section{Introduction}

\def\spw{0.24}

\begin{figure}[t]
  \centering
  \includegraphics[width=\spw\linewidth]{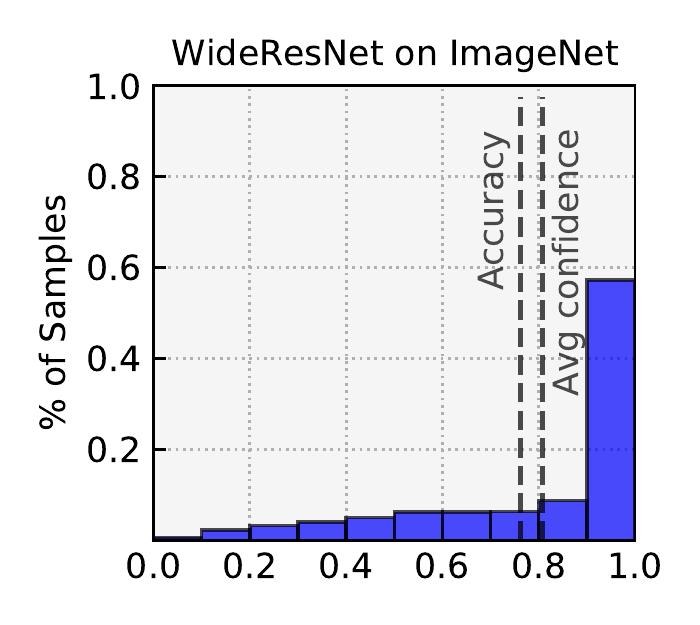}
  \includegraphics[width=\spw\linewidth]{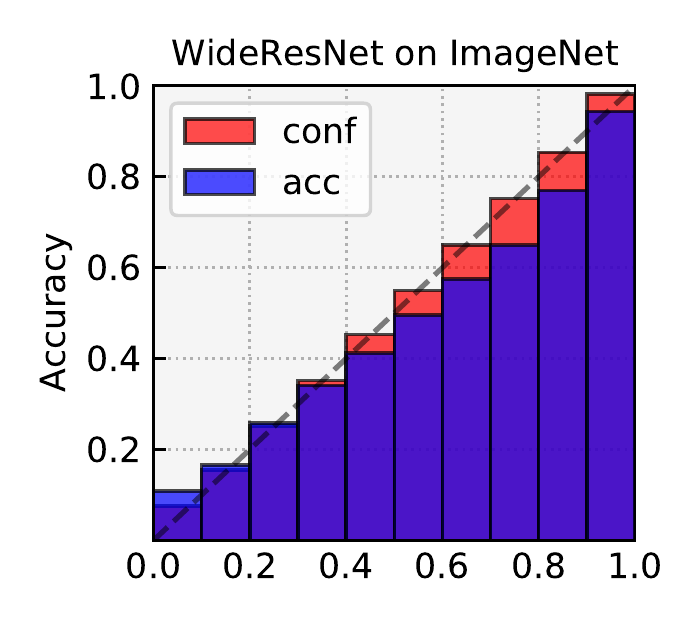}
  \includegraphics[width=\spw\linewidth]{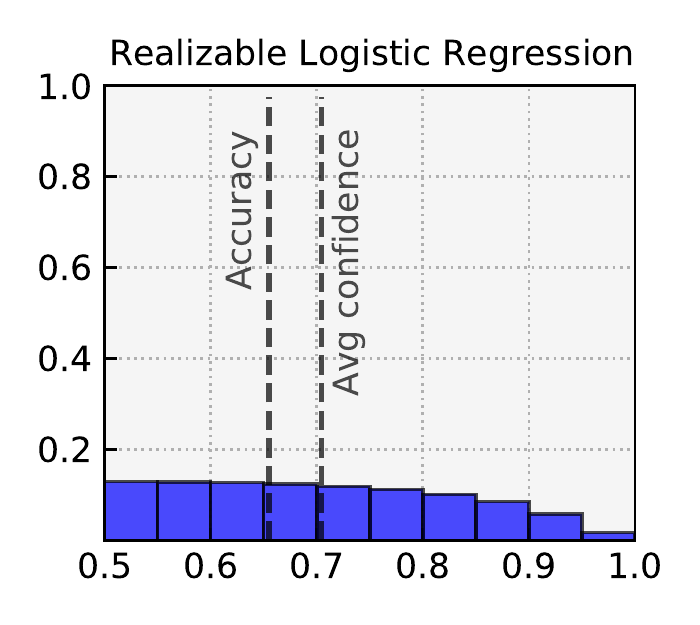}
  \includegraphics[width=\spw\linewidth]{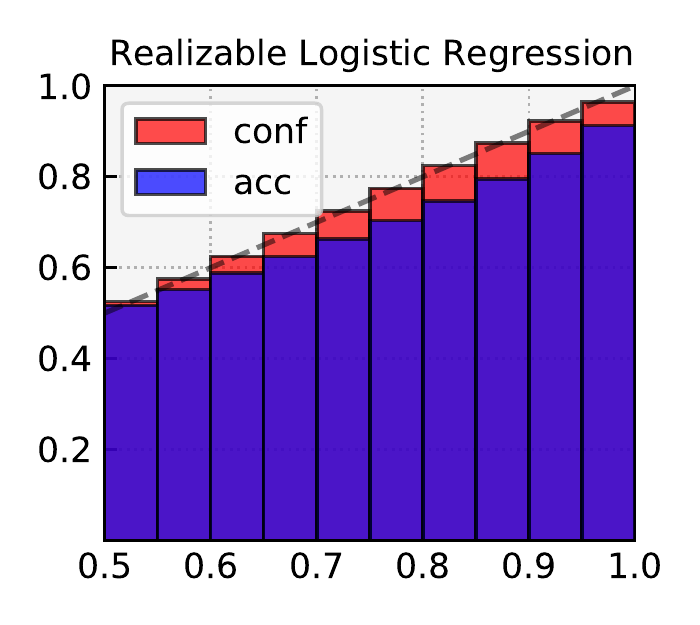}
  \caption{\small Reliability diagrams for calibration: Over-parametrized deep network vs. well-specified, under-parametrized logistic regression. The $x$-axes denote the confidences (predicted top probabilities) of the models. Left \& Middle Left: WideResNet-50-2 on ImageNet. Middle Right \& Right: Binary logistic regression on simulated data with $n=2000$ and $d=100$.}
  \label{figure:fig1}
\end{figure}


Modern machine learning models such as deep neural networks with high accuracy tend to be miscalibrated: The predicted top probability (\emph{confidence}) does not reflect the actual accuracy of the model, and tends to be \emph{over-confident}. For example, a WideResNet 32 on CIFAR100 has on average a predicted top probability of $87\%$, while the actual test accuracy is only $72\%$~\citep{guo2017calibration}. As the confidence is often comprehended as an estimate of the true accuracy, such over-confidence could be dangerous, especially in risk-sensitive domains such as medical AI~\citep{begoli2019need}, self-driving cars~\citep{michelmore2018evaluating}, and so on. To address this issue, there is a growing line of research on improving the calibration of models, by either performing \emph{recalibration} of well-trained models to adjust the confidence scores~\citep{platt1999probabilistic,zadrozny2001obtaining,naeini2015obtaining,guo2017calibration}, or by averaging the predictions over multiple models to make the confidence scores more accurate~\citep{lakshminarayanan2016simple,gal2016dropout}. These methods in general can reduce the over-confidence and improve the calibration of the model, while preserving (or even improving) the model's accuracy~\citep{ovadia2019can}.

Despite these progresses, the more fundamental question of \emph{why} such over-confidence happens for vanillaly trained models remains not satisfactorily understood. One common understanding is that over-confidence is a result of \emph{over-parametrization}: Models such as deep neural networks are large enough to memorize the entire training dataset, and are encouraged to magnify its weights and maximize the confidence so as to minimize the training loss~\citep{mukhoti2020calibrating}. \citet{guo2017calibration}~also observed that increasing the depth and width makes the over-confident more severe, even when this improves the accuracy. However, so far it is unclear whether over-parametrization is the only reason, or whether there are other intrinsic reasons leading to over-confidence.


In this paper, we show that over-confidence is not just a result of over-parametrization and is more inherent. We conduct a precise theoretical study on the calibration in binary classification problems. Our main result shows that \emph{standard logistic regression is also over-confident}, even in the well-specified, under-parametrized scenario where the model is correct (data generated from a linear logistic model), and there is abundant data (number of samples $n$ much greater than number of parameters $d$).

Figure~\ref{figure:fig1} illustrates our main finding via simulation: Similar to an over-parametrized neural network, the empirical risk minimizer of logistic regression is also over-confident at all confidence levels. Note that these two models have rather different behaviors in terms of the distribution of confidences, yet their over-confidence behaviors are similar.

Our contributions are summarized as follows:
\begin{itemize}
\item We show that \emph{well-specified logistic regression is inherently over-confident}: Conditioned on the model predicting $p>0.5$, the actual probability of the label being one is lower by an amount of $\Theta(d/n)$, in the limit of $n,d\to\infty$ proportionally and $n/d$ is large (Section~\ref{section:logistic}). In other words, the calibration error is always in the over-confident direction. We also show that the overall Calibration Error (CE) of the logistic model is $\Theta(d/n)$ in this limiting regime.

\item We identify sufficient conditions for over- and under-confidence in general binary classification problems, where the data is generated from an arbitrary nonlinear activation, and we solve a well-specified empirical risk minimization (ERM) problem with a suitable loss function (Section~\ref{section:general}). Our conditions imply that any symmetric, monotone activation $\sigma:\R\to[0,1]$ that is \emph{concave} at all $z> 0$ will yield a classifier that is over-confident at any confidence level.

\item Another perhaps surprising implication is that \emph{over-confidence is not universal}: We prove that there exists an activation function for which under-confidence can happen for a certain range of confidence levels.
 
\item We perform simulation and real data experiments to test our theory (Section~\ref{section:experiments}). Our experiments suggest that the over-confidence of logistic regression happens broadly in a variety of under-parametrized settings, within or beyond our theory's assumptions. We also verify that under-confidence can indeed happen in simulations with the activation function constructed above.

\item On the technical end, our analysis develops a precise understanding of the high-dimensional proportional limit of ERM in the \emph{sufficient data} regime ($n/d$ is large) by rigorously establishing the first-order behavior of the solution to the characterizing system of nonlinear equations (Section~\ref{section:proof-overview}), which may be of broader interest.
\end{itemize}


\subsection{Related work}

\paragraph{Algorithms for model calibration}
Practitioners have observed and dealt with the over-confidence of logistic regression long ago. \emph{Recalibration algorithms} fix this by adjusting the output of a well-trained model, and dates back to the classical methods of Platt scaling~\citep{platt1999probabilistic}, histogram binning~\citep{zadrozny2001obtaining} and isotonic regression~\citep{zadrozny2002transforming}.~\citet{platt1999probabilistic} also uses a particular kind of label smoothing as a way of mitigating the over-confidence in logistic regression.
\citet{guo2017calibration} show that temperature scaling, a simple method that learns a rescaling factor for the logits, is a competitive method for calibrating neural networks. A number of recent recalibration methods further improve the performances over these approaches~\citep{kull2017beta,kull2019beyond,ding2020local,rahimi2020intra,zhang2020mix}.

Another line of work improves calibration by aggregating the probabilisitic predictions over multiple models, using either an ensemble of models~\citep{lakshminarayanan2016simple,malinin2019ensemble,wen2020batchensemble,tran2020hydra}, or randomized predictions such as Bayesian neural networks~\citep{gal2016dropout,gal2017concrete,maddox2019simple,dusenberry2020efficient}. Finally, there are techniques for improving the calibration of a single neural network during training~\citep{thulasidasan2019mixup,mukhoti2020calibrating,liu2020simple}.

\paragraph{Theoretical analysis of calibration}
\citet{kumar2019verified} show that continuous rescaling methods such as temperature scaling is less calibrated than reported, and proposed a method that combines temperature scaling and histogram binning. \citet{gupta2020distribution} study the relationship between calibration and other notions of uncertainty such as confidence intervals. \citet{shabat2020sample,jung2020moment} study the sample complexity of estimating the multicalibration error (group calibration). A related theoretical result to ours is \citep{liu2019implicit} which shows that the calibration error of any classifier is upper bounded by its square root excess logistic loss over the Bayes classifier. This result can be translated to a $O(\sqrt{d/n})$ upper bound for well-specified logistic regression, whereas our main result implies $\Theta(d/n)$ calibration error in our high-dimensional limiting regime (with input distribution assumptions).


\paragraph{High-dimensional behaviors of empirical risk minimization}

There is a rapidly growing literature on limiting characterizations of convex optimization-based estimators in the $n \propto d$ regime \citep{donoho2009message, bayati2011dynamics, el2013robust, karoui2013asymptotic, stojnic2013framework, thrampoulidis2015regularized, thrampoulidis2018precise, mai2019large, sur2019modern, candes2020phase}.
Our analysis builds on the characterization for unregularized convex risk minimization problems (including logistic regression) derived in \citet{sur2019modern}.

%% file: Sections-arxiv/prelim.tex
\section{Preliminaries}
In this paper we consider binary classification problems, where we observe $n$ data points $\set{(\xb_i,y_i)}_{i=1}^n\simiid P$ for some distribution $P$ on $\R^d\times \set{0,1}$.

\subsection{Calibration}
Let $\hat{f}:\R^d\to[0,1]$ be a (probabilistic) classifier. $\hat{f}$ is said to be perfectly calibrated if $\P(Y=1|\hat{f}(\Xb)=p)=p$ for all $p\in[0,1]$, that is, the actual probability of $Y=1$ conditioned on $\what{f}$ predicting $p$ is exactly $p$. In reality, we cannot hope for obtaining perfect calibration, and would rather desire ways of measuring the calibration error.

A standard metric is the Calibration Error (CE), which measures the difference between the prediction and the conditional mean of $Y$ given the prediction~\citep{guo2017calibration}:
\begin{align}
  \label{equation:ce}
  \CE(\hat{f}) \defeq \E_{(\Xb,Y)\sim P} \brac{ \abs{\hat{f}(\Xb) - \E[Y \mid \hat{f}(\Xb)] } } .
\end{align}
Notably, CE is the population (unbinned) version of the Expected Calibration Error (ECE), a commonly used calibration metric in recent  work~\citep{naeini2015obtaining,guo2017calibration,ovadia2019can,nixon2019measuring}.

In this paper, we consider the \emph{calibration error of $\what{f}$ at level $p$}:
\begin{align}
  \label{equation:deltapcal}
  \deltapcal(\hat{f}) \defeq p - \P_{(X,Y)\sim P}\paren{ Y=1 \mid \hat{f}(\Xb)=p }
\end{align}
for all $p\in(0,1)$. Note that $\deltapcal(\hat{f})$ is the quantity inside the expectation in~\eqref{equation:ce}, and provides a more fine-grained characterization of the calibration error by specifying which $p$ we are interested in.

\paragraph{Over-confidence and under-confidence}
The \emph{confidence} of $\hat{f}$ at $\xb$ is the predicted top probability, i.e. $\max\{\hat{f}(\xb), 1-\hat{f}(\xb)\}$ for binary problems. In particular, when $\hat{f}(\xb)>0.5$, the confidence is equal to $\hat{f}(\xb)$. We say that the model is \emph{over-confident} when the confidence is higher than the actual accuracy: For example, when the model predicts $\hat{f}(\xb)=0.9$, but we have $\E[Y|\hat{f}(\xb)=0.9]=0.8$, then $\hat{f}$ is over-confident at level $p=0.9$. Note that in this case the calibration error at level $0.9$ is positive: $\Delta_{0.9}^{\sf cal}(\hat{f})=0.1>0$. In other words, over- or under-confidence is determined by the {\bf sign} of the calibration error $\deltapcal(\hat{f})$ in definition~\eqref{equation:deltapcal}:
\begin{tcolorbox}
For any $p\in(0.5,1)$:
\begin{itemize}
\item $\deltapcal(\hat{f})>0$: $\hat{f}$ is {\bf over-confident} at level $p$;
\item $\deltapcal(\hat{f})<0$: $\hat{f}$ is {\bf under-confident} at level $p$.
\end{itemize}
\end{tcolorbox}
We remark that we only state results for $p>0.5$ in this paper; all the results also hold for $p\in(0,0.5)$ by symmetry.

\paragraph{Extension to multi-class problems}
In our experiments we also consider multi-class classification problems, for which
there is a standard generalization of definitions~\eqref{equation:deltapcal} and~\eqref{equation:ce}~\citep{guo2017calibration}: Given a multi-class predictor $\hat{F}:\R^d\to \Delta_K$ where $K\ge 2$ is the number of classes, we replace $Y$ with the indicator of correct prediction: $\indic{Y = \argmax_k \hat{F}(x)_k}$, and replace $\hat{f}(x)$ with the confidence $\max_k \hat{F}(x)_k$. Thus the calibration error of $\hat{F}$ at level $p\in[1/K, 1]$ is
$
\deltapcal(\hat{F})
\defeq p - \P\paren{ Y = \argmax_k \hat{F}(\Xb)_k \mid \max_k \hat{F}(\Xb)_k = p  }
$.

\subsection{Model and data distribution}
We consider the following data distribution where $\Xb$ is standard Gaussian and $Y|\Xb$ follows a binary linear model with activation function $\sigma:\R\to[0,1]$:
\begin{align}
  \label{equation:gaussian-realizable}
  P:~~~\Xb\sim\normal(0, \Ib_d),~~~\P(Y=1 \mid \Xb=\xb)=\sigma(\wb_\star^\top\xb),
\end{align}
where $\wb_\star\in\R^d$ is the ground truth coefficient vector. (This is also known as generalized linear models with link function $\sigma$~\citep{mccullagh2018generalized}). We make the Gaussian input assumption as our analysis requires a precise limiting calculation; however, our real data experiments in Section~\ref{section:experiments-real} suggest that the implications of our theory may hold more broadly without such distributional assumptions.



\paragraph{Realizable logistic regression}
Our primary focus is \emph{realizable logistic regression}, in which $\sigma(z)=\frac{1}{1+e^{-z}}$ is the logistic (sigmoid) activation, and we solve the unregularized ERM (empirical risk minimization) problem
\begin{equation}
  \label{equation:logistic-erm}
  \begin{aligned}
    \hat{\wb} = \argmin_{\wb} \hat{R}_n(\wb) \defeq \frac{1}{n}\sum_{i=1}^n \brac{ \log(1 + \exp(\wb^\top \xb_i)) - y_i\wb^\top \xb_i }.
  \end{aligned}
\end{equation}
Let $R(\wb)\defeq \E[\hat{R}_n(\wb)]$ denote the expected (population) risk. It is a classical result that $\argmin_\wb R(\wb)=\wb_\star$, i.e. logistic regression is well-specified when data comes from the logistic model~\citep{hastie2009elements}.


\paragraph{Extension to general activations}
We also consider generalizations where $\sigma$ is a general monotone activation function, and we wish to learn a linear classifier $\hat{\wb}$ that is close to $\wb_\star$. In this case, we consider solving the general ERM
\begin{align}
  \label{equation:erm}
  {\rm minimize} ~~\hat{R}_n(\wb) \defeq \frac{1}{n}\sum_{i=1}^n \rho(\wb^\top\xb_i) - y_i\wb^\top\xb_i,
\end{align}
where $\rho:\R\to\R$ is a loss function. Let $R(\wb)\defeq \E[\hat{R}_n(\wb)]$ denote the expected (population) risk.

To make sure the problem is well-specified, we choose $\rho$ to be the (integrated) convex loss associated with $\sigma$: $\rho(z) = \int_0^z \sigma(u)du + C$ for some constant $C$; in other words $\rho'(z)=\sigma(z)$. It is known that for such a choice of $\rho$ we have $\argmin_\wb R(\wb) = \wb_\star$~\citep{kakade2011efficient}. (For completeness we also provide a proof in Appendix~\ref{appendix:general-convex-erm}.)


We require the following assumption on the activation function $\sigma$ along with the loss function $\rho$, which only requires the activation to be smooth along with some basic properties, such as monotonicity and symmetry around $0$.
\begin{assumption}[Smooth activation]
  \label{assumption:activation_new}
  The loss function $\rho:\R\to\R$ is strictly convex and four-times continuously differentiable with uniformly bounded \{1, 2, 3, 4\}-th derivatives. The activation function $\sigma=\rho'$ is strictly increasing, and satisfies $\sigma(0)=1/2$, $\lim_{z\to-\infty}\sigma(z)=0$, $\lim_{z\to\infty}\sigma(z)=1$, and $\sigma'(z)=\sigma'(-z)> 0$ for all $z\in\R$. 
\end{assumption}






%% file: Sections-arxiv/logistic.tex
\section{Logistic regression is over-confident}
\label{section:logistic}

As a warm-up, consider running unregularized (linear) logistic regression in the over-parametrized setting where $n<d$ and the data is separable. In this case, it is known that the (ERM) solution to the logistic regression~\eqref{equation:logistic-erm} does not exist~\citep{albert1984existence,candes2020phase}; the gradient descent path will also diverge to infinity norm~\citep{soudry2018implicit}. Using an approximate solution $\hat{\wb}$ with a high norm will cause the learned classifier $\sigma(\hat{\wb}^\top\xb)$ to be nearly a step function (outputs are close to either $0$ or $1$). Such classifiers are clearly over-confident whenever the true conditional distribution $Y|\Xb$ is not approximately deterministic.


We are now ready to present our main result, which states that even in the most vanilla setting (well-specified, under-parametrized), logistic regression is still over-confident.
\begin{theorem}[Well-specified logistic regression is over-confident]
  \label{theorem:logistic}
  Consider the classifier $\hat{f}(\xb)=\sigma(\hat{\wb}^\top\xb)$ obtained from logistic regression~\eqref{equation:logistic-erm}, where the data is generated from the logistic model~\eqref{equation:gaussian-realizable}. Then we have the following. 
  \begin{itemize}
  \item In the limit of $n,d\to\infty$\footnote{We assume $\norm{\wb_\star}$ is the same for all $(n,d)$.} and $d/n\to\kappa$, where $\kappa\in (0, \kappa_0]$ for some constant $\kappa_0>0$ (which only depends on $\norm{\wb_\star}$), for any $p\in (0.5, 1)$, almost surely, we have
  \begin{align*}
    \deltapcal(\hat{f}) \to C_{p, \kappa} ~~~\textrm{for some}~C_{p, \kappa} > 0.
  \end{align*}
  In words, logistic regression gives inherently over-confident estimates of the actual probabilities. 
  \item We have, for small enough $\kappa>0$,
  \[
  C_{p, \kappa} = C_p \cdot \kappa + o(\kappa). 
  \]
  In words, as the sample size $n/d = 1/\kappa$ becomes large, the over-confidence effect becomes weaker. The scaling of this over-confidence effect is roughly $C_p\cdot d/n$. 
  \end{itemize}
\end{theorem}


\paragraph{Over-confidence is inherent for logistic regression}
Theorem~\ref{theorem:logistic} considers the under-parametrized setting, as we allow $d/n=\kappa$ to be any \emph{small} value, thus the sample size $n$ can be arbitrarily higher than the dimension $d$. It thus suggests that over-confidence of logistic regression a rather fundamental property, and challenges the common belief that over-confidence mostly comes from over-parametrization. Furthermore, even though $\deltapcal(\hat{f})$ becomes smaller as the sample size increases ($\kappa$ becomes lower), Theorem~\ref{theorem:logistic} still asserts the sign of $\deltapcal(\hat{f})$ being always positive in the proportional limit of $n,d\to\infty$, $d/n\to\kappa$. This result perhaps unveils another source of over-confidence in real-world machine learning models beyond linear logistic models. 

Furthermore, Theorem~\ref{theorem:logistic} shows that logistic regression is over-confident at all $p\in(0.5,1)$. This suggests that the over-confidence in every confidence bin, as an empirical observation in well-trained neural networks~\citep{guo2017calibration}, holds for logistic regression as well and is not unique to large over-parametrized models. 


\paragraph{Regularization; comparison with classical asymptotics}
We remark that our result only holds for \emph{unregularized} logistic regression, while it is known that various regularization can improve calibration~\citep{gal2016dropout,thulasidasan2019mixup}.
Indeed, in our model, applying regularization (e.g. an L2 regularizer) will in general reduce the calibration error, as long as the regularization reduces the norm of $\hat{\wb}$ and does not hurt its correlation with $\wb_\star$ too much.
However, we intentionally focus on unregularized logistic regression which resembles practical setups such as neural networks in the memorizing regime. We also note that, in general, the best regularization strength for the optimal accuracy and the optimal calibration may be different.

We also briefly remark that our setting of $d,n\to\infty$, $d/n\to\infty$ is different from classical asymptotic statistics (which considers fixed $d$ and $n\to\infty$)~\citep{van2000asymptotic}. Classical asymptotics would imply $\sqrt{n}\deltapcal(\hat{f})\gotod\normal(0,V^2)$ for some $V^2$, and thus $\deltapcal(\hat{f})$ has about equal chance to be positive or negative; in contrast, we show that $\deltapcal(\hat{f})$ has a positive bias in the proportional limit, a regime arguably more realistic than classical asymptotics.



\paragraph{CE of logistic regression}
Theorem~\ref{theorem:logistic} further implies a result on the calibration error (CE) of logistic regression.


\begin{corollary}[Asymptotics of calibration error]
  \label{corollary:logistic-ce}
  In the same setting as Theorem~\ref{theorem:logistic}, as $d,n\to\infty$, $d/n\to\kappa$, the $\CE$ of the logistic regression solution $\hat{f}$ satisfies
  \begin{align*}
    \quad \CE(\what{f}) \to C_\kappa, 
  \end{align*}
  almost surely, where for small enough $\kappa$ we have $C_\kappa   = C\kappa + o(\kappa)$ for some absolute constant $C>0$. 
\end{corollary}
Corollary~\ref{corollary:logistic-ce} implies that, in the limiting regime, the CE of logistic regression is $O(\kappa)=O(d/n)$. This improves over the results of~\citet{liu2019implicit} in certain aspects.
First, \citet[Corollary 2.4]{liu2019implicit}~showed that the CE of any classifier is bounded by the square root excess logistic loss over the Bayes classifier. This implies the CE of well-specified logistic regression is bounded by $\sqrt{d/n}$. Here we show the CE has a better rate $\Theta(d/n)$ at small $d/n$ in our limiting regime\footnote{We remark that Corollary~\ref{corollary:logistic-ce} does not readily imply a $\Theta(d/n)$ result in the non-asymptotic setting. However, we believe a similar result (with additional terms such as $1/\sqrt{n}$) holds and can be established via a more refined analysis.}.
Second, our Theorem~\ref{theorem:logistic} determines the sign of the calibration error (confidence $>$ accuracy), which is not implied by their results.





The proof of Corollary~\ref{corollary:logistic-ce} follows directly from Theorem~\ref{theorem:logistic} by integrating $\deltapcal(\hat{f})$ over $p\in(0, 1)$ (with $p$ distributed as $\hat{f}(\xb)$ for $\xb \sim P$). The proof can be found in Appendix~\ref{appendix:proof-logistic-ce}.


\subsection{Proof sketch of Theorem~\ref{theorem:logistic}}
\label{section:proof-sketch-logistic}

We now provide a high-level overview of the proof of Theorem~\ref{theorem:logistic}. A more detailed overview of the most technical steps is deferred to Section~\ref{section:proof-overview}, and the full proofs can be found in Appendix~\ref{appendix:proportional} \&~\ref{appendix:proofs}.


\paragraph{Closed-form expression for calibration error}
Recall that
\begin{align*}
  \deltapcal(\hat{f}) = p - \E_{\xb}\brac{ \sigma(\wb_\star^\top \xb) \mid \sigma(\hat{\wb}^\top \xb) = p}.
\end{align*}
As $\xb$ is standard Gaussian, the conditional distribution of $\xb | \sigma(\hat{\wb}^\top \xb)=p$ can be characterized precisely in terms of the projection of $\xb$ onto the direction $\hat{\wb}$ and its orthogonal complement subspace. Standard calculation then yields the closed form expression
\begin{equation}
  \label{equation:deltapcal-expression}
  \begin{aligned}
    \deltapcal(\hat{f}) = p - \E_Z\brac{ \sigma\paren{ \frac{\norm{\wb_\star}}{\norm{\hat{\wb}}}\cos\hat{\theta}\cdot \sigma^{-1}(p) + \sin\hat{\theta}\norm{\wb_\star}Z} },
  \end{aligned}
\end{equation}
where $\cos\hat{\theta}=\frac{\hat{\wb}^\top\wb_\star}{\norm{\hat{\wb}}\norm{\wb_\star}}$ is the angle between $\hat{\wb}$ and $\wb_\star$, and $Z\sim \normal(0,1)$. (See Lemma~\ref{lemma:calibration-error} for the detailed statement and proof.)





\paragraph{Concentration of $\hat{\wb}$}
In the second step, we apply results from recent advances in high-dimensional convex risk minimization~\citep{sur2019modern,taheri2020fundamental} to show that $\hat{\wb}$ concentrates around fixed values in the high-dimensional limit, in terms of its norm and cosine angle with $\wb_\star$. These results show that, in the limit of $d,n\to\infty$ and $d/n\to\kappa$, the following concentration happens almost surely:
\begin{equation}
  \label{equation:concentration-rc}
  \begin{aligned}
    & \norm{\hat{\wb}} \to \normbar = \normbar(\kappa, \norm{\wb_\star}), \\
    & \cos\hat{\theta} \to \cosbar = \cosbar(\kappa, \norm{\wb_\star}),
  \end{aligned}
\end{equation}
Above, $\normbar$ and $\cosbar$ are determined by the solutions of a system of nonlinear equations with three variables $(\alpha, \sigma, \lambda)$ (see Section~\ref{section:proof-overview} and Theorem~\ref{theorem:limit} for the formal statement).

The concentration directly implies that $\deltapcal(\hat{f})$ converges to the following limiting calibration error (Corollary~\ref{corollary:concentration-deltapcal}):
\begin{equation}
  \label{equation:limiting-deltapcal}
  \begin{aligned}
    \deltapcal(\hat{f}) \to C_{p,\kappa} \defeq p - \E_Z\brac{ \sigma\paren{ \frac{\norm{\wb_\star}}{\normbar}\cosbar \cdot \sigma^{-1}(p) + \sqrt{1-\cosbar^2}\norm{\wb_\star}Z }}.
  \end{aligned}
\end{equation}
This expression hints on potential sources of over- or under-confidence: (1) $\normbar$ and $\cosbar$ will affect the ``multiplier'' $\norm{\wb_\star}\cosbar/\normbar$ in front of $\sigma^{-1}(p)$, drifting the expectation away from $p$; (2) $\cosbar$ also affects the expectation over the $\sqrt{1-\cosbar^2}\norm{\wb_\star}Z$ term.
This term itself has mean zero, but can affect the overall expectation through the nonlinear activation function $\sigma$.






\paragraph{Calculating the limiting calibration error}
The final part, as a technical crux of the proof, calculates the limiting calibration error~\eqref{equation:limiting-deltapcal} by precisely analyzing the interplay between the concentration values $\normbar$, $\cosbar$ and the activation function $\sigma$. This is achieved by a novel analysis on the solutions of the aformentioned system of equations at small $\kappa$. In particular, we show that $C_{p,\kappa}= C_p\kappa+o(\kappa)$ for small $\kappa$, and $C_p>0$ is positive, thereby establishing Theorem~\ref{theorem:logistic}. We present a more detailed description of this analysis in Section~\ref{section:proof-overview}.

%% file: Sections-arxiv/general.tex
\section{Over-confidence is not universal}
\label{section:general}
It is natural to ask---based on Theorem~\ref{theorem:logistic}---whether over-confidence is true in other well-specified problems as well, or is due to some specific property about logistic regression. This section makes steps towards this by looking at the generalized problem~\eqref{equation:erm} where $\sigma$ is an arbitrary activation and we solve the corresponding convex ERM.


Our main result in this section is the following characterization of sufficient conditions for whether over- or under-confidence happens in the general convex ERM~\eqref{equation:erm}. The proof of this result can be found in Appendix~\ref{appendix:proof-general}.

\begin{theorem}[Sufficient conditions for over- and under-confidence]
  \label{theorem:general}
  In the same setting as Theorem~\ref{theorem:logistic} except that the activation function $\sigma$ is general and satisfies Assumption~\ref{assumption:activation_new}, let $\hat{f}(\xb)=\sigma(\hat{\wb}^\top\xb)$ be the classifier obtained from the convex ERM~\eqref{equation:erm}. We have simultaneously for any $p\in(0.5, 1)$ that, almost surely in the limit of $d,n\to\infty$, $d/n\to\kappa$,
  \begin{align}
    \label{equation:general-limit}
    \deltapcal(\hat{f}) \to C_{p,\kappa}(\sigma) = C_p(\sigma)\kappa + o(\kappa). 
  \end{align}
  Further, we have the following sufficient conditions for the sign of $C_{p}(\sigma)$: For any $p\in(0.5, 1)$,
  \begin{enumerate}[(a)]
  \item If $\sigma$ is concave at $\sigma^{-1}(p)$, i.e., 
    \begin{align}
      \sigma''(\sigma^{-1}(p)) \le 0,
    \end{align}
    then $C_{p}(\sigma)>0$, and $\hat{f}$ is over-confident at this $p$ for all sufficiently small $\kappa$.
  \item Conversely, if
    \begin{align}
      & \E_{Q_1\sim \normal(0, \norm{\wb_\star}^2)} \brac{ Q_1\sigma''(Q_1) } > 0,~~~{\rm and}~\label{equation:suff-uc-1} \\
      & \sigma''(\sigma^{-1}(p)) - 2\sigma'(\sigma^{-1}(p))\cdot \sigma^{-1}(p)/\norm{\wb_\star}^2 > 0,~\label{equation:suff-uc-2}
    \end{align}
    then $C_p(\sigma)<0$, and $\hat{f}$ is under-confident at this $p$ for all sufficiently small $\kappa$.
  \end{enumerate}
\end{theorem}

\paragraph{Interpretations}
Theorem~\ref{theorem:general} suggests that the \emph{curvature} of the activation function $\sigma$ is critical for determining its over- or under-confidence. We parse these sufficient conditions as follows:
\begin{itemize}
\item \emph{Concavity of $\sigma(z)|_{z>0}$ implies over-confidence}: By part (a), at any $p$ where $\sigma''(\sigma ^{-1}(p))\ge 0$, $\hat{f}$ will be over-confident at that $p$. Moreover, any $\sigma$ that is concave on the entire positive part ($\sigma''(z)\le 0$ for all $z>0$) will result in over-confident at every $p>0.5$. This strictly generalizes Theorem~\ref{theorem:logistic}, and suggests that over-confidence is a common mode, as 
  any $\sigma$ that is monotone and bounded must have some concave regions on the positive part. 
\item \emph{Under-confidence is possible but cannot hold at every $p$}: Part (b) suggests that under-confidence may be possible, provided we design $\sigma$ that is sufficiently \emph{convex} at $\sigma^{-1}(p)$ (to counteract the other term in~\eqref{equation:suff-uc-2}), and that additional condition~\eqref{equation:suff-uc-1} holds. However, as $\sigma''(z)>0$ cannot happen for all $z>0$, under-confidence cannot happen at every $p\in(0.5, 1)$. 
\end{itemize}



Is the sufficient condition for under-confidence in Theorem~\ref{theorem:general}(b) indeed possible? We give an affirmative answer.
\begin{corollary}[Under-confidence can happen]
  \label{corollary:underconfidence}
  There exists an activation function $\sigma$ satisfying Assumption~\ref{assumption:activation_new}, such that $C_{p}(\sigma)<0$ for some $p\in (0.5, 1)$ and $\norm{\wb_\star}>0$. At these $p$, the convex ERM~\eqref{equation:erm} is under-confident in the limit of $d/n\to\kappa$ for all small $\kappa$. 
\end{corollary}
The activation we find in Corollary~\ref{corollary:underconfidence} is very close to the following activation function (up to minor tweaks in order to satisfy Assumption~\ref{assumption:activation_new}):
\begin{equation}
  \label{equation:cos-activation}
  \sigma_{\uc}(z) = \left\{
  \begin{aligned}
    & 0, & z < -2\pi, \\
    & \frac{1}{2} + \frac{1}{4\pi}(z - \sin z), & |z| \le 2\pi, \\
    & 1, & z > 2\pi.
  \end{aligned}
  \right.
\end{equation}
(See Figure~\ref{figure:sigma} for the plot of this activation function.) The unique feature about this $\sigma_{\uc}$ is that, unlike the logistic activation, this function is convex at all small values of $z>0$. This leads to both the convexity condition~\eqref{equation:suff-uc-2} as well as the expectation condition~\eqref{equation:suff-uc-1} (which roughly requires the positive part in the expectation of $Q_1\sigma''(Q_1)$ to supercede the negative part).

To the best of our knowledge, this is the first known case of under-confidence for a well-specified classification problem, though we remark this under-confidence effect is weak and restricted to only a small region of $p$ (see Figure~\ref{figure:cal-052} for simulation results using this activation).






%% file: Sections-arxiv/experiments.tex

\begin{figure}[t]
  \centering
  \begin{minipage}{0.32\textwidth}
    \centering
    \subcaption{\small Illustration of $\sigma_{\uc}$}\label{figure:sigma}
    \includegraphics[width=\textwidth]{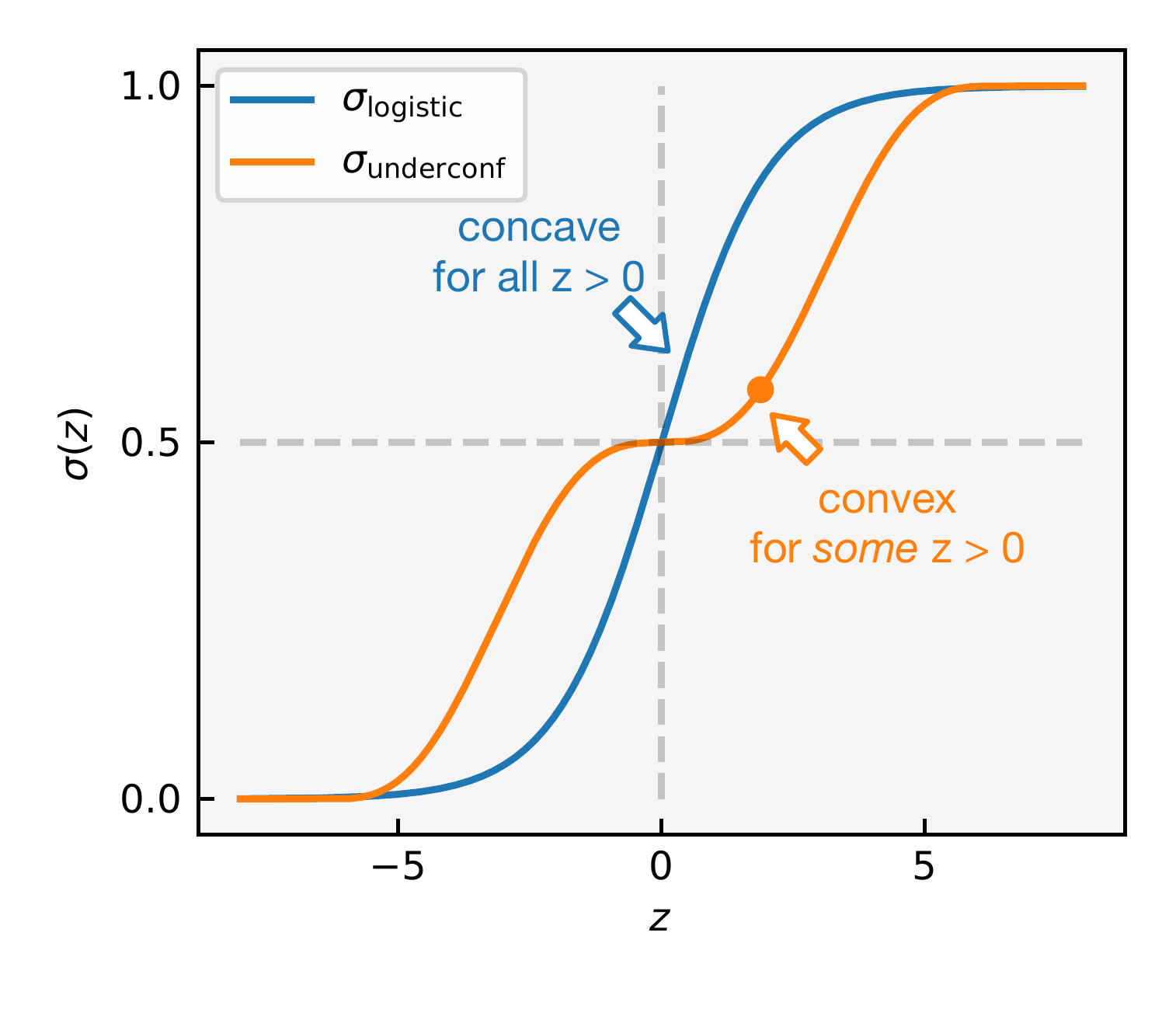}
  \end{minipage}
  \begin{minipage}{0.32\textwidth}
    \centering
    \subcaption{\small Calibration of logistic regression}\label{figure:cal-099}
    \includegraphics[width=\textwidth]{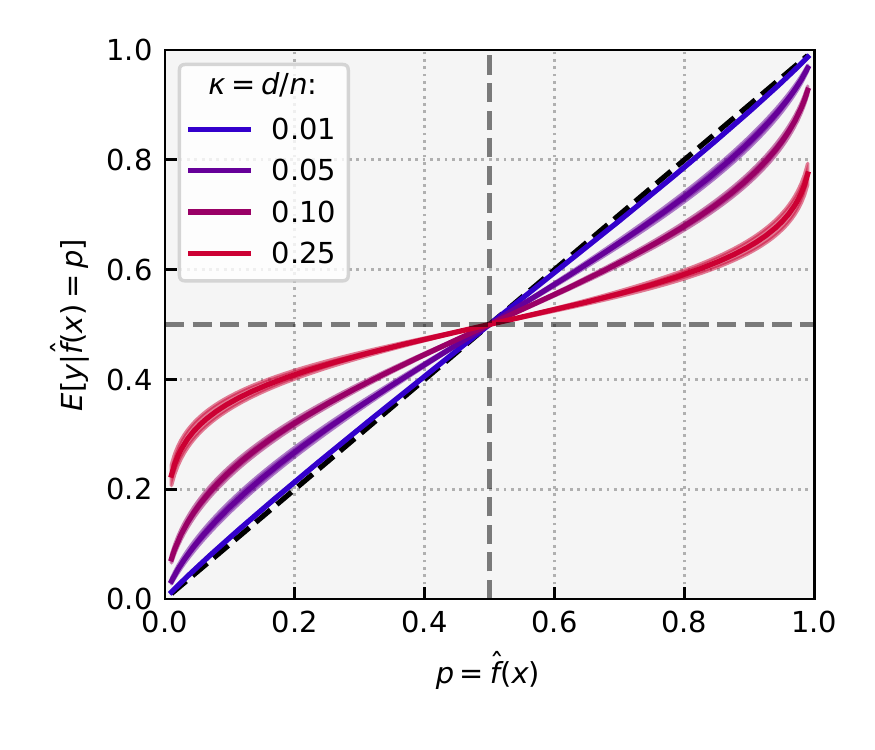}
  \end{minipage}
    \begin{minipage}{0.32\textwidth}
      \centering
      \subcaption{\small Calibration with $\sigma_{\uc}$}\label{figure:cal-052}
      \includegraphics[width=\textwidth]{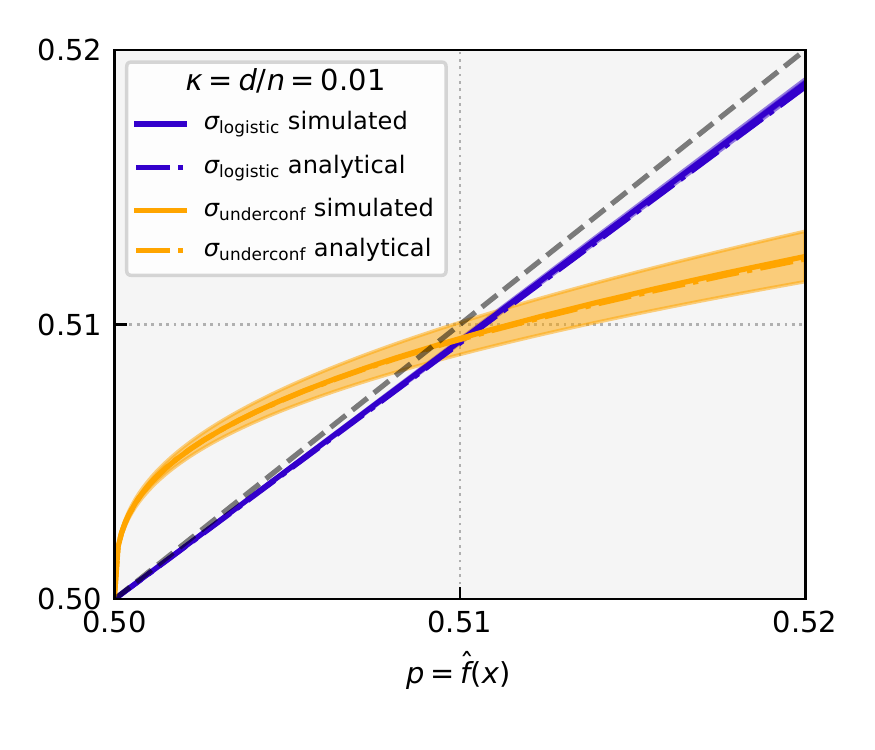}
    \end{minipage}
    \vspace{-1em}
  \caption{
    Binary classification simulations on realizable data.
    {\bf (a)} Illustration of the activation function $\sigma_{\uc}$ constructed in Corollary~\ref{corollary:underconfidence} (cf.~\eqref{equation:cos-activation}), against the logistic (sigmoid) activation $\sigma_{\logistic}$.
    {\bf (b)} Calibration curves for simulated logistic regression, with $d=100$ and $d/n\in\set{0.01, 0.05, 0.1, 0.25}$. {\bf Logistic regression is over-confident} (prediction is higher than actual probability when prediction$>0.5$) at all $d/n$.
    {\bf (c)} Zoomed-in calibration curves for simulated realizable ERM with the $\sigma_{\uc}$ activation. In contrast to logistic regression, {\bf $\sigma_{\uc}$ leads to under-confidence} for $p\in (0.5, 0.51)$, verifying our Theorem~\ref{theorem:general} and Corollary~\ref{corollary:underconfidence}. Here ``analytical'' refers to our theoretical prediction $p - C_{p,\kappa}(\sigma)$ from Theorem~\ref{theorem:general}.
    (b)(c): Shaded areas are one-std error bars over $5$ runs.
  }
  \label{figure:simulation}
\end{figure}

\section{Experiments}
\label{section:experiments}

\subsection{Simulations}
\label{section:simulations}

We test our theories via simulations on well-specified under-parametrized logistic regression, as well as general convex ERM with the under-confident activation $\sigma_{\uc}$~\eqref{equation:cos-activation}.

For both activations, we generate data $\set{(\xb_i, y_i)}_{i=1}^n$ from the realizable model~\eqref{equation:gaussian-realizable}, where we fix $d=100$, $\norm{\wb_\star}=1$, and vary $d/n\in\set{0.01,0.05,0.10,0.25}$. For each $(d,n)$, we generate 5 problem instances, solve the ERM problem on each instance, and plot the ``calibration curves'' (where the $x$-axis is $p$ and $y$-axis is the average probability given the prediction: $\E[y|\hat{f}^{(i)}(\xb)=p] = p - \deltapcal(\hat{f}^{(i)})$), visualizing their mean and one-standard-deviation error bar. Notice that by the closed-form expression~\eqref{equation:deltapcal-expression}, we are able to compute $\deltapcal(\hat{f})$ exactly (using Gaussian integration) without needing to introduce a test set.

In addition to the simulated calibration curves, we also plot the limiting calibration curve suggested by Theorem~\ref{theorem:logistic} \&~\ref{theorem:general}, in which we compute the concentration values $\normbar,\cosbar$ analytically by solving its defining equations (cf. Appendix~\ref{appendix:equations}), 
and plug these values into the closed-form expression~\eqref{equation:limiting-deltapcal}. This yields a curve of $p$ against $p-C_{p,\kappa}(\sigma)$, which we compare against our simulated curves.


\paragraph{Results}
Figure~\ref{figure:simulation} shows the results of our simulations. We find logistic regression indeed yields over-confident calibration curves (Figure~\ref{figure:cal-099}): $\E[y|\hat{f}(\xb)=p]=p-\deltapcal(\hat{f})$ is less than $p$ for $p>0.5$ (and greater than $p$ for $p<0.5$). Further, notice that the gap $\deltapcal$ increases as we increase $\kappa$. This agrees with our intuition that over-confidence is more severe when $d/n$ increases (effective sample size gets lower), and further suggests that the conclusion of our theory holds more broadly than its assumptions: $\kappa$ can be as large as $\kappa=0.25$ and $d$ can be as low as $100$, both being realistic values for modeling practice.

We also find that the under-confidence shown in Corollary~\ref{corollary:underconfidence} does show up in the simulations: With the activation $\sigma_{\uc}$, $\E[y|\hat{f}(\xb)=p]$ is \emph{higher} than $p$ for $p\in(0.5, 0.51)$, although this range of $p$ is fairly narrow (Figure~\ref{figure:cal-052}).

Finally, we observe that our theoretical prediction $C_{p,\kappa}$ closely matches the simulation: the analytical calibration curve $p-C_{p,\kappa}(\sigma)$ and the mean simulated curve are almost identical for both activations, which further confirms our theory even at a realistic $d=100$.







\begin{figure*}[t]
  \centering
  \begin{minipage}{0.49\textwidth}
    \centering
    \subcaption{True labels}\label{figure:cifar-true}
    \vspace{-.7em}
    \includegraphics[width=0.49\textwidth]{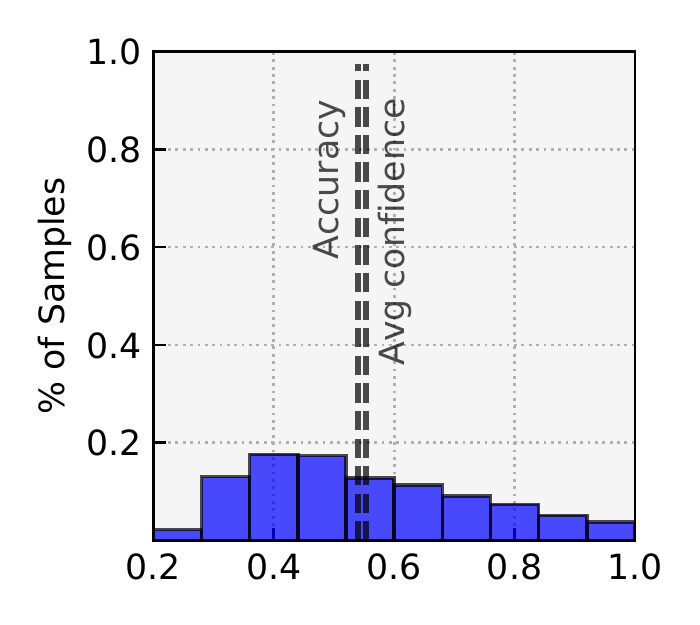}
    \includegraphics[width=0.49\textwidth]{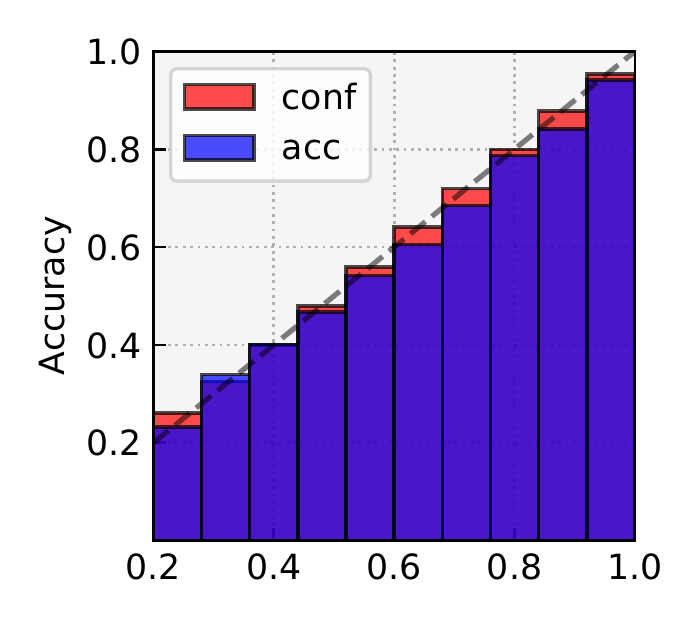}
  \end{minipage}
  \begin{minipage}{0.49\textwidth}
    \centering
    \subcaption{Pseudo labels (realizable)}\label{figure:cifar-pseudo}
    \vspace{-.7em}
    \includegraphics[width=0.49\textwidth]{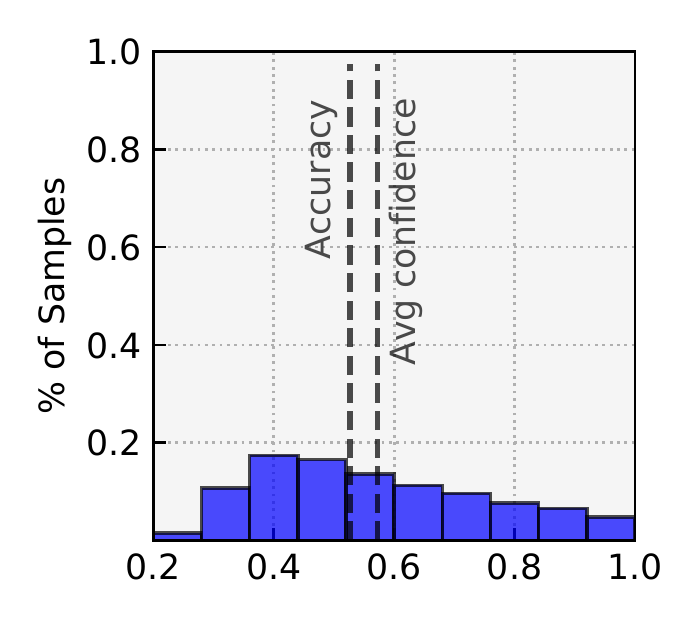}
    \includegraphics[width=0.49\textwidth]{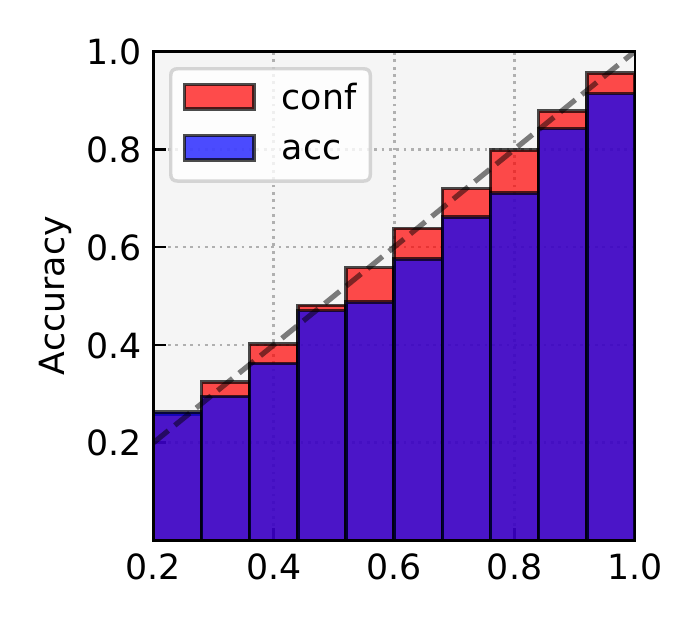}
  \end{minipage}
  \vspace{-1em}
  \caption{ Calibration of multi-class logistic regression on CIFAR10's first 5-classes. The $x$-axes denote the confidences (predicted top probabilities) of the models.
    {\bf (a)(b)}: Left: Confidence distribution across bins; Right: Average confidence against average accuracy within each bin (right); both evaluated on the test set.
    {\bf (a)} Logistic regression on the true labels.
    {\bf (b)} Logistic regression on pseudo-labels generated from the fitted logistic model (realizable setting) from step (a).
    Observe that {\bf over-confidence happens for both the pseudo-labels generated from a multi-class logistic model, and the true labels.}}
  \label{figure:cifar10}
\end{figure*}

\subsection{CIFAR10 with pseudo labels}
\label{section:experiments-real}

We further test the generality of our theory beyond the Gaussian input assumption and the binary classification setting. We run multi-class logistic regression on the first 5 classes of CIFAR10, which contains $n=25000$ training images and 5000 test images, and each image has $d=3072$ features. We perform logistic regression on two kinds of labels:
\begin{itemize}
\item The true label $y^{\sf true}\in\set{0,1,2,3,4}$.
\item The pseudo-label $y^{\sf pseudo}\in\set{0,1,2,3,4}$ generated as follows: After fitting the logistic classifier $\hat{\Wb}\in\R^{3072\times 5}$ on the true labels, we generate pseudo-labels $y_i^{\sf pseudo}$ from the multi-class logistic (softmax) model
  \begin{align*}
    \P\paren{y_i^{\sf pseudo} = k \mid \xb_i} = \frac{\exp(\hat{\Wb}_k^\top \xb_i)}{\sum_{k'} \exp(\hat{\Wb}_{k'}^\top \xb_i)}.
  \end{align*}
\end{itemize}
The motivation for the pseudo-labels is to construct a well-specified problem (labels do come from a linear softmax model) and remove the potential effect of model-misspecification with the true labels. Note that this problem is still in the under-parametrized setting as $d<n$.

As the exact conditioning $\hat{f}(\xb)=p$ is no longer computable on finite data, we compute the average confidence and accuracy on the test set using binning (10 equally spaced confidence bins in $[0.2,1.0]$), similar as in the standard practice for evaluating the ECE~\citep{guo2017calibration}. Additional experimental details are provided in Appendix~\ref{appendix:cifar10-details}.

\paragraph{Results}
We find that logistic regression on (the 5-class subset of) CIFAR10 is over-confident on both the pseudo-labels and true labels (Figure~\ref{figure:cifar10}).
A closer look reveals that the over-confidence is more severe on the pseudo-labels than the true labels, yet both tasks exhibit a reasonable level of over-confidence (especially in the high confidence bins).
This suggests our result that logistic regression is inherently over-confident may hold more broadly for other under-parametrized problems without strong assumptions on the input distribution, or even when the labels are not necessarily realizable by the model.





%% file: Sections-arxiv/proof-sketch.tex
\section{Overview of analysis}
\label{section:proof-overview}

This section provides an overview of the two novel proof steps for our results in Section~\ref{section:logistic} \&~\ref{section:general}: (1) Characterization of the high-dimensional limit (concentration value) of logistic regression~\eqref{equation:logistic-erm} and the general convex ERM~\eqref{equation:erm} at small $\kappa=d/n$. (2) Determining the sign of the limiting calibration error based on the above characterization, filling in the (abbreviated) last part of the proof sketch in Section~\ref{section:proof-sketch-logistic}.


\subsection{Local linear analysis at small $\kappa$}
\label{section:local-linear-analysis}

Let $\gamma\defeq \norm{\wb_\star}$. By the results of~\citet{sur2019modern}, the values $\normbar,\cosbar$ in~\eqref{equation:concentration-rc} have the form $\normbar=\sqrt{\alpha_\star^2 + \kappa\sigma_\star^2}$ and $\cosbar=(1+\kappa\sigma_\star^2/\alpha_\star^2\gamma^2)^{-1/2}$, where $(\alpha_\star,\sigma_\star, \lambda_\star)$ are the solutions to the following system of nonlinear equations in three variables $(\alpha, \sigma, \lambda)$:

\begin{equation*}
  \label{equation:system_new}
  \left\{
  \begin{aligned}
    & \sigma^2 = \frac{1}{\kappa^2} \E\brac{ 2\rho'(Q_1) \lambda^2\rho'(\prox_{\lambda\rho}(Q_2))^2}, \\
    & 0 = \E\brac{ \rho'(Q_1) Q_1 \lambda\rho'(\prox_{\lambda\rho}(Q_2)) },\\
    & 1-\kappa = \E\brac{
      2\rho'(Q_1) \big/ (1 + \lambda\rho''(\prox_{\lambda\rho}(Q_2)))
      }.
  \end{aligned}
  \right.
\end{equation*}

Above, $(Q_1,Q_2)$ has a bivariate normal distribution with covariance depending on $(\alpha, \sigma, \kappa, \gamma)$, and $\prox$ is the prox operator. (See Theorem~\ref{theorem:limit} and Appendix~\ref{appendix:equations} for a formal statement.)
These solutions are guaranteed to uniquely exist for small enough $\kappa$. However, they are only implicitly defined without closed-form expressions for these solutions, which prohibits us from analyzing their behaviors.

We overcome this issue by performing a local analysis of the solutions at small $\kappa$. We prove that, for small enough $\kappa$, we have the local linear approximation
\begin{align*}
  & \alpha_\star = \alpha_\star(\kappa) = 1 + \baralpha_0 \kappa + O(\kappa^2), \\
  & \sigma_\star = \sigma_\star^2(\kappa) = \barsigma_0^2 + O(\kappa), \\
  & \lambda_\star = \lambda_\star(\kappa) = \barlambda_0\kappa + O(\kappa^2),
\end{align*}
with closed-form expressions for $(\baralpha_0, \barsigma_0, \barlambda_0)$. For example, we have $\barsigma_0^2 = \E[\rho'(Q_1)\rho'(-Q_1)]/\paren{\E[\rho''(Q_1)]}^2$ where $Q_1\sim\normal(0,\gamma^2)$. (See Lemma~\ref{lem:small_kappa_asymptotics} for the formal statement.) These approximations imply similar approximations for $\normbar,\cosbar$, which allows us to analyze the behavior of the limiting calibration error~\eqref{equation:limiting-deltapcal} locally at small $\kappa$.




\subsection{Determining sign of the limiting calibration error}
Towards proving Theorem~\ref{theorem:general} \&~\ref{theorem:logistic}, it remains for us to derive the sufficient conditions for the sign of $C_{p,\kappa}(\sigma)$. Using the above local linear approximation for $(\normbar,\cosbar)$ and performing first-order calculus, we obtain
\begin{align*}
  \lim_{\kappa\to 0} \frac{C_{p,\kappa}(\sigma)}{\kappa} & = \sigma'(\sigma^{-1}(p)) \cdot \sigma^{-1}(p) \cdot \paren{\baralpha_0 + \barsigma_0^2/\gamma^2} - \frac{1}{2}\sigma''(\sigma^{-1}(p)) \cdot \barsigma_0^2.
\end{align*}
(Lemma~\ref{lemma:cpkappa-small-kappa}). We prove that $\baralpha_0 + \barsigma_0^2/\gamma^2>0$ always holds regardless of the activation,$\gamma$, and $p$. This implies that, as long as $\sigma''(\sigma^{-1}(p))\le 0$, the right-hand side in the equation above is positive. This gives part (a) of Theorem~\ref{theorem:general}. On the other hand, if $\sigma''(\sigma^{-1}(p))>2\sigma'(\sigma^{-1}(p))\sigma^{-1}(p)/\gamma^2$ and $\baralpha_0<0$, then the right-hand side above is negative. These are exactly the sufficient conditions we required in part (b) of Theorem~\ref{theorem:general}.




%% file: Sections-arxiv/conclusion.tex
\section{Conclusion}
This paper provides a precise theoretical study of the calibration error of logistic regression and a class of general binary classification problems. We show that logistic regression is inherently over-confident by $\Theta(d/n)$ when $n/d$ is large, and establish sufficient conditions for the over- or under-confidence of unregularized ERM for general binary classification. Our results reveal that (1) Over-confidence is not just a result of over-parametrization; (2) Over-confidence is a common mode but not universal. We believe our work opens up a number of future questions, such as the interplay between calibration and model training (or regularization), or theoretical studies of calibration on nonlinear models.







%% file: Sections-arxiv/acknowledgment.tex
\section*{Acknowledgment}
We thank Yuxin Chen for the insightful discussions on the system of nonlinear equations. We thank all anonymous reviewers for the many helpful feedback on the conference version of this paper.

%% file: Sections-arxiv/tools.tex
\section{Technical tools}
\label{appendix:tools}

\subsection{The prox operator}

For any loss function $\rho$ that satisfies Assumption \ref{assumption:activation_new}, for any $\lambda > 0$, we define its proximal mapping operator via
\begin{equation}
\prox_{\lambda \rho}(z) \equiv \arg \min_{t \in \R} \Big\{ \lambda \rho(t) + \frac{1}{2} (t - z)^2 \Big\}. 
\end{equation}

\begin{lemma}[Properties of the Proximal mapping operator]\label{lem:proximal}
For any loss function $\rho$ that satisfies Assumption \ref{assumption:activation_new}, its proximal mapping operator $\prox_{\lambda \rho}(z)$ is differentiable with respect to $z \in \R$ and $\lambda > 0$. Moreover, we have
\[
\begin{aligned}
\frac{\de}{\de z} \prox_{\lambda \rho}(z) =&~ \frac{1}{1 + \lambda \rho''( \prox_{\lambda \rho}(z))}, \\
\frac{\de}{\de \lambda} \prox_{\lambda \rho}(z) =&~ - \frac{\rho'(\prox_{\lambda \rho}(z)) }{ 1 + \lambda \rho''( \prox_{\lambda \rho}(z)) } . 
\end{aligned}
\]
\end{lemma}

\begin{proof}[Proof of Lemma \ref{lem:proximal}]

Denote $P(\lambda, z) = \prox_{\lambda \rho}(z)$. For any loss function $\rho$ that satisfies Assumption \ref{assumption:activation_new}, the proximal mapping operator $P = P(\lambda, z)$ satisfies the following equation
\[
\lambda \rho'(P(\lambda, z)) + P(\lambda, z) - z = 0. 
\]

Taking derivative with respect to $z$, we have 
\[
\lambda \rho''(P(\lambda, z)) \partial_z P(\lambda, z) +  \partial_z P(\lambda, z) - 1 = 0,  
\]
which gives 
\[
\partial_z P(\lambda, z)  = \frac{1}{1 + \lambda \rho''(P(\lambda, z))}. 
\]
Taking derivative with respect to $\lambda$, we have 
\[
\rho'(P(\lambda, z)) + \lambda \rho''(P(\lambda, z)) \partial_\lambda P(\lambda, z)  +  \partial_\lambda P(\lambda, z) = 0, 
\]
which gives
\[ 
\partial_\lambda P(\lambda, z) = - \frac{\rho'(P(\lambda, z)) }{ 1 + \lambda \rho''( P(\lambda, z)) } . 
\]
This proves the lemma. 
\end{proof}

\subsection{Implicit function theorem}

We state the standard implicit function theorem in the following.
\begin{lemma}[Implicit function theorem]\label{lem:implicit_function}
Let $\bF(\bp, \kappa):\R^{d}\times \R_{\ge 0}\to \R^d$ be a continuously differentiable vector-valued function on $\ball(\bp_0, \eps) \times [0, \bar \kappa_0)$ for some $\bar \kappa_0 > 0$. Suppose $\bF(\bp_0, 0) = 0$ and 
\[
\sigma_{\min}(\nabla_\bp F(\bp_0, 0)) > 0. 
\]
Then there exists a constant $\kappa_0 > 0$ and a continuous differentiable path $\bp_\star(t) \in \ball(\bp_0, \eps)$, such that 
\[
\bF(\bp_\star(\kappa), \kappa) = 0, ~~~ \forall \kappa \in [0, \kappa_0).
\]
\end{lemma}

\subsection{Consistency of the convex ERM for general activation}
\label{appendix:general-convex-erm}
Recall 
\begin{align*}
  R(\wb) = \E\brac{ \rho(\wb^\top\xb) - y\wb^\top\xb}
\end{align*}
and $\E\brac{y|\xb} = \sigma(\wb_\star^\top\xb)$. If $\rho'=\sigma$, then we have
\begin{align*}
  \grad R(\wb) = \E\brac{ \rho'(\wb^\top\xb)\xb - y\xb } = \E\brac{ \paren{\sigma(\wb^\top\xb) - \sigma(\wb_{\star}^\top\xb)} \xb }.
\end{align*}
This shows that $\grad R(\wb_\star)=0$, which further by the convexity of $\rho$ implies that $\wb_\star \in \argmin_{\wb} R(\wb)$.
\qed

%% file: Sections-arxiv/closed-form.tex
\section{Closed form expression for calibration error}
\label{appendix:closed-form}

\begin{lemma}[Closed-form expression for calibration error]
  \label{lemma:calibration-error}
  Assume $(\Xb, Y)$ follows the binary linear model~\eqref{equation:gaussian-realizable} with true coefficient $\wb_\star\in\R^d$, and the activation function $\sigma$ satisfies Assumption~\ref{assumption:activation_new}. For any $\wb\in\R^d$, and $f_\wb(\xb)\defeq \sigma(\wb^\top\xb)$, we have for any $p\in(0,1)$ that
  \begin{equation}
    \begin{aligned}
      \deltapcal(f_\wb) = p - \E_{Z\sim \normal(0,1)}\brac{ \sigma\paren{ \frac{\norm{\wb_\star}}{\norm{\wb}}\cos\theta\cdot \sigma^{-1}(p) + \sin\theta\norm{\wb_\star}Z} },
    \end{aligned}
  \end{equation}
  where $\theta=\angle(\wb, \wb_\star)$ is the angle between $\wb$ and $\wb_\star$.
\end{lemma}
\begin{proof}
  Recall by definition~\eqref{equation:deltapcal} that
  \begin{align*}
    \deltapcal(f_\wb) = p - \E_{\xb, y} \brac{ y \mid f_\wb(\xb) = p } = p - \E_{\xb} \brac{ \sigma(\wb_\star^\top\xb) \mid \sigma(\wb^\top\xb) = p }.
  \end{align*}
  As $\xb\sim\normal(0,1)$, the conditional distribution of $\xb | \sigma(\wb^\top \xb)=p$ can be characterized precisely: Under this distribution we have $\wb^\top \xb = \sigma^{-1}(p)$ and $\Vb^\top\xb\sim \normal(0, \Ib_{d-1})$, where $\Vb\in\R^{d\times (d-1)}$ is the orthogonal complement subspace of $\wb$. Therefore, conditioned on $\sigma(\wb^\top \xb)=p$, we have
  \begin{align*}
    & \quad \wb_\star^\top\xb = \wb_\star^\top \frac{1}{\norm{\wb}^2} \wb\wb^\top\xb + \wb_\star^\top \Vb^\top\xb \\
    & \eqnd \frac{\wb_\star^\top\wb}{\norm{\wb}^2} \sigma^{-1}(p) + \norm{\wb_\star}\sin\angle(\wb_\star, \wb) \cdot Z = \frac{\norm{\wb_\star}}{\norm{\wb}} \cos\theta \cdot \sigma^{-1}(p) + \norm{\wb}_\star\sin\theta\cdot  Z,
  \end{align*}
  where $\eqnd$ denotes equal in distribution, and $Z\sim\normal(0,1)$. This representation of the condition distribution yields the desired expression for $\deltapcal(f_\wb)$.
\end{proof}

%% file: Sections-arxiv/proof-proportional.tex
\section{Analysis of convex ERM in the high-dimensional limit}
\label{appendix:proportional}

\subsection{Concentration to a system of equations}
\label{appendix:equations}




We define the following system of equations in three variables $(\alpha, \sigma, \lambda)\in\R^3$, with two parameters $\kappa>0$ and $\gamma>0$:
\begin{subequations}
  \label{equation:system_new}
  \begin{empheq}[left=\empheqlbrace]{align}
    & \sigma^2 = \frac{1}{\kappa^2} \E\brac{ 2\rho'(Q_1) \lambda^2\rho'(\prox_{\lambda\rho}(Q_2))^2}, \label{equation:1_new} \\
    & 0 = \E\brac{ \rho'(Q_1) Q_1 \lambda\rho'(\prox_{\lambda\rho}(Q_2)) }, \label{equation:2_new} \\
    & 1-\kappa = \E\brac{  \frac{2\rho'(Q_1)}{1 + \lambda\rho''(\prox_{\lambda\rho}(Q_2)) }}, \label{equation:3_new}
  \end{empheq}
\end{subequations}
where $(Q_1,Q_2)$ follows a joint normal distribution with mean $\bzero$ and covariance matrix
\begin{align*}
  \bSigma =
  \begin{bmatrix}
    \gamma^2 & -\alpha\gamma^2 \\
    -\alpha\gamma^2 & \alpha^2\gamma^2 + \kappa\sigma^2
  \end{bmatrix}.
\end{align*}
Let
\begin{align}
  \label{equation:solutions}
  (\alpha_\star, \sigma_\star, \lambda_\star) = (\alpha_\star(\kappa), \beta_\star(\kappa), \lambda_\star(\kappa))
\end{align}
denote the solution to~\eqref{equation:system_new} whenever the solution exists and is unique (dropping dependence on $\gamma$ for notational simplicity).

It is recently shown that the solutions $(\alpha_\star, \sigma_\star, \lambda_\star)$ is closely connected to the high-dimensional limit of the convex ERM~\eqref{equation:erm}. We state this in the following result.

\begin{theorem}[Concentration of the ERM~\eqref{equation:erm}; restatement of Theorem 2,~\citep{sur2019modern}]
  \label{theorem:limit}
  Let Assumption~\ref{assumption:activation_new} hold. For any $\gamma=\norm{\wb_\star}$, there exists a $\kappa_0=\kappa_0(\gamma)>0$ such that solution $(\alpha_\star,\sigma_\star,\lambda_\star)$ defined in~\eqref{equation:solutions} uniquely exists in the positive region $\Omega=\set{(\alpha, \sigma, \lambda):\alpha>0,\sigma>0,\lambda>0}$ for any $\kappa\in(0,\kappa_0]$.

  Further, let $\what{\wb}$ denote the solution to the convex ERM~\eqref{equation:erm}, and assume the data is generated from the binary linear model~\eqref{equation:gaussian-realizable} with $\norm{\wb_\star}=\gamma$ fixed. As $d,n\to\infty$ with $d/n\to\kappa$ where $\kappa\in(0,\kappa_0]$ is a fixed constant, we have almost surely that $\hat{\wb}$ exists, and concentrates in the sense that
  \begin{equation}
    \label{equation:concentration-erm}
    \begin{aligned}
      & \norm{\hat{\wb}} \to \normbar = \normbar(\kappa, \gamma) \defeq \sqrt{\alpha_\star^2\gamma^2 + \kappa\sigma_\star^2}, \\
      & \cos\hat{\theta} \defeq \frac{\hat{\wb}^\top\wb_\star}{\norm{\hat{\wb}} \norm{\wb_\star}} \to \cosbar = \cosbar(\kappa, \gamma) \defeq \frac{1}{\sqrt{1 + \kappa\sigma_\star^2/\alpha_\star^2\gamma^2}}.
    \end{aligned}
  \end{equation}
\end{theorem}

Combining Lemma~\ref{lemma:calibration-error} with our the concentration result in Theorem~\ref{theorem:limit} directly implies the following
\begin{corollary}[Concentration of the calibration error]
  \label{corollary:concentration-deltapcal}
  In the same setting as Theorem~\ref{theorem:limit}, the calibration error $\deltapcal(\hat{f})$ (of the convex ERM) converges almost surely to the following limit as $d,n\to\infty$, $d/n\to\kappa$:
  \begin{align}
    \label{equation:concentration-deltapcal}
    \deltapcal(\hat{f}) \to  C_{p,\kappa}(\sigma) \defeq p -
    \E_{Z\sim\normal(0,1)}\brac{ \sigma\paren{ \frac{\norm{\wb_\star}}{\normbar}\cosbar \cdot \sigma^{-1}(p) + \sqrt{1-\cosbar^2}\norm{\wb_\star}Z }},
  \end{align}
  where $\normbar,\cosbar$ are defined in~\eqref{equation:concentration-erm}.
\end{corollary}
\begin{proof}
  This is a direct consequence of the concentration result of Theorem~\ref{theorem:limit}. By Lemma~\ref{lemma:calibration-error} we have for the ERM learned classifier $\hat{f}(\xb)=\sigma(\hat{\wb}^\top\xb)$ that
  \begin{align*}
    \deltapcal(\hat{f}) = p - \E_{Z\sim \normal(0,1)}\brac{ \sigma\paren{ \frac{\norm{\wb_\star}}{\norm{\hat{\wb}}}\cos\hat{\theta}\cdot \sigma^{-1}(p) + \sin\hat{\theta}\norm{\wb_\star}Z} },
  \end{align*}
  where $\cos\hat{\theta} = \wb_\star^\top\hat{\wb}/\norm{\wb_\star}\norm{\hat{\wb}}$ is the cosine similarity between $\wb_\star$ and $\hat{\wb}$. Now, by Theorem~\ref{theorem:limit}, we have $\norm{\hat{\wb}}\to \normbar$, $\cos\hat{\theta}\to\cosbar$, and $\sin\hat{\theta}\to \sqrt{1-\cosbar^2}$ with probability one as $n,d\to\infty$, $d/n\to\kappa$. (Note that the sign of $\sin\hat{\theta}$ does not make a difference here as $Z\sim\normal(0,1)\eqnd -Z$.) Therefore, the random variable inside the expectation in the above converges to the random variable
  \begin{align*}
    \sigma\paren{ \frac{\norm{\wb_\star}}{\normbar}\cosbar \cdot \sigma^{-1}(p) + \sqrt{1-\cosbar^2}\norm{\wb_\star}Z }
  \end{align*}
  with probability one. Applying the bounded convergence theorem (since $|\sigma(\cdot)|\le 1$) yields that $\deltapcal(\hat{f})$ converges to
  \begin{align*}
    C_{p,\kappa}(\sigma) \defeq p - \E_{Z\sim\normal(0,1)}\brac{ \sigma\paren{ \frac{\norm{\wb_\star}}{\normbar}\cosbar \cdot \sigma^{-1}(p) + \sqrt{1-\cosbar^2}\norm{\wb_\star}Z }},
  \end{align*}
  as desired.
\end{proof}


\subsection{Analysis of solution at small $\kappa$}
We now analyze the solution $(\alpha_\star, \beta_\star, \kappa_\star)$ when $\kappa$ is small. Throughout this section, we let $Q_2=-\alpha Q_1 + \sqrt{\kappa}\sigma Z$, so that $Z\sim\normal(0,1)$ and is independent of $Q_1$.  Then we have $(Q_1, Q_2) \sim \cN(\bzero, \bSigma)$, where $\bSigma = [\gamma^2, -\alpha \gamma^2; -\alpha \gamma^2, \alpha^2 \gamma^2 + \kappa \sigma^2]$. 

For any $(\baralpha, \barsigma, \barlambda, \kappa) \in \overline \Omega \times (0, \kappa_0)$ where $\overline \Omega = \{ (\baralpha, \barsigma, \barlambda): \baralpha > -1/\kappa, \barsigma > 0, \barlambda > 0\}$, we consider the following equivalent version of the system of equations~\eqref{equation:system_new}, defined via
\begin{equation}\label{eqn:def_bF}
\begin{aligned}
F_1(\baralpha, \barsigma, \barlambda, \kappa) =&~ \barsigma^2 - \barlambda^2 \E[2 \rho'(Q_1) \rho'(\prox_{\kappa \barlambda \rho}(- (1 + \kappa \baralpha) Q_1 + \sqrt{\kappa} \barsigma Z))^2], \\
F_2(\baralpha, \barsigma, \barlambda, \kappa) =&~ \kappa^{-1}\E[\rho'(Q_1) Q_1 \rho'(\prox_{\kappa \barlambda \rho}(- (1 + \kappa \baralpha) Q_1 + \sqrt{\kappa} \barsigma Z))], \\
F_3(\baralpha, \barsigma, \barlambda, \kappa) =&~ \kappa^{-1} \{1 - \kappa - \E[2 \rho'(Q_1) / [1 + \kappa \barlambda \rho''(\prox_{\kappa \barlambda\rho}(- (1 + \kappa \baralpha) Q_1 + \sqrt{\kappa} \barsigma Z))] ]\}.  
\end{aligned}
\end{equation}
We further define $\bp = (\baralpha, \barsigma, \barlambda)$ as a shorthand for the three variables. For $(\bp, \kappa) \in \overline \Omega \times (0, \kappa_0)$, we let
\begin{equation}\label{eqn:def_bbF}
\bF(\bp, \kappa) = (F_1(\bp, \kappa), F_2(\bp, \kappa), F_3(\bp, \kappa)). 
\end{equation}
Equation~\eqref{eqn:def_bbF} and the system~\eqref{equation:system_new} are equivalent up to a change of variables: For any fixed $\kappa$, any solution $(\alpha_\star, \sigma_\star, \lambda_\star)$ of Eq. (\ref{equation:system_new}) yields a solution $((\alpha_\star - 1) / \kappa, \sigma_\star, \kappa \lambda_\star, \kappa)$ of $\bF(\bp, \kappa) = \bzero$ where $\bF$ is defined as in Eq. (\ref{eqn:def_bbF}), and vice versa. As a consequence, Theorem~\ref{theorem:limit} implies that, for any $\kappa \in (0, \kappa_0)$, there exists a unique solution $\bp(\kappa)$ of $\bF(\bp, \kappa) = 0$ in the corresponding region $\overline \Omega = \{ (\baralpha, \barsigma, \barlambda): \baralpha > -1/\kappa, \barsigma > 0, \barlambda > 0\}$. 

Now we define
\begin{subequations}
  \label{equation:bar0}
  \begin{empheq}[]{align}
    \baralpha_0 \defeq&~ -\frac{\E\brac{Q_1\rho'''(Q_1)} \cdot \E\brac{\rho'(Q_1)\rho'(-Q_1)}}{ 2\E\brac{Q_1^2\rho''(Q_1)} \cdot \E\brac{\rho''(Q_1)}^2 }, \label{equation:baralpha0}\\
    \barsigma_0^2 \defeq&~ \frac{\E[\rho'(Q_1)\rho'(-Q_1)]}{\paren{\E[\rho''(Q_1)]}^2 } > 0, \label{equation:barsigma0} \\
    \barlambda_0 \defeq&~ \frac{1}{\E[\rho''(Q_1)]} > 0, \label{equation:barlambda0}
  \end{empheq}
\end{subequations}
where the expectation is taken with respect to $Q_1 \sim \normal(0, \gamma^2)$. 

\begin{lemma}[Local linear analysis of solutions at small $\kappa$]
  \label{lem:small_kappa_asymptotics}
  In the same setting as Theorem~\ref{theorem:limit},
  the solutions $(\alpha_\star, \sigma_\star, \lambda_\star)$ defined in~\eqref{equation:solutions} satisfy
  \begin{subequations}
    \label{equation:limit}
  \begin{empheq}[]{align}
    & \alpha_\star(\kappa) = 1 + \baralpha_0 \kappa + o(\kappa), \label{equation:limit-1} \\
    & \sigma_\star(\kappa) = \barsigma_0 + o(1), \label{equation:limit-2} \\
    & \lambda_\star(\kappa) = \barlambda_0 \kappa + o(\kappa), \label{equation:limit-3}
  \end{empheq}
\end{subequations}
where $(\baralpha_0, \barsigma_0, \barlambda_0)$ are defined in~\eqref{equation:bar0}, and $o(\cdot)$ is the standard small-o notation ($o(\kappa)/\kappa\to 0$ as $\kappa\to 0$).
\end{lemma}
Lemma~\ref{lem:small_kappa_asymptotics} formalizes the arguments we sketched in Section~\ref{section:local-linear-analysis}. (We note that Section~\ref{section:local-linear-analysis} stated the error terms in terms of the slightly stronger $O(\kappa^2)$ instead of $o(\kappa)$, yet $o(\kappa)$ as stated above is sufficient for our subsequent results.) The rest of this section is devoted to proving Lemma~\ref{lem:small_kappa_asymptotics}.

\subsection{Proof of Lemma \ref{lem:small_kappa_asymptotics}}

We take $\bp_0 = (\baralpha_0, \barsigma_0, \barlambda_0)$, where $\baralpha_0, \barsigma_0, \barlambda_0$ are defined in~\eqref{equation:bar0}. Observe that, in order to prove~\eqref{equation:limit}, it suffices to prove that under the change of variables
\begin{align*}
  (\baralpha(\kappa), \barsigma(\kappa), \barlambda(\kappa)) \defeq \paren{ \frac{\alpha_\star(\kappa)-1}{\kappa}, \sigma_\star(\kappa), \frac{\lambda_\star(\kappa)}{\kappa}},
\end{align*}
the new set of solutions obey
\begin{align*}
  & \baralpha(\kappa) = \baralpha_0 + o(1), \\
  & \barsigma(\kappa) = \barsigma_0 + o(1), \\
  & \barlambda(\kappa) = \barlambda_0 + o(1).
\end{align*}
In other words, letting $\bp(\kappa)=(\baralpha(\kappa), \barsigma(\kappa), \barlambda(\kappa))$ denote the solution to the transformed equation $\bF(\bp, \kappa)=0$, we wish to prove that $\bp(\kappa)\to \bp_0$.  We achieve this through a continuity analysis of the function $\bF$ and the implicit function theorem. The key intermediate results are the following two auxiliary lemmas, whose proofs are deferred to Section~\ref{appendix:proof-jacobian} and \ref{appendix:proof_lem_F_kappa_derivative} respectively.

\begin{lemma}\label{lem:Jacobian}
Let Assumption \ref{assumption:activation_new} hold. Let $\bF$ be as defined in Eq. (\ref{eqn:def_bbF}). Then for any $\eps$ such that $\ball(\bp_0, 2 \eps) \subseteq \overline \Omega = \{ (\baralpha, \barsigma, \barlambda): \baralpha > - 1/\kappa, \barsigma > 0, \barlambda>0 \}$, there exists a continuous matrix function $\bJ: \ball(\bp_0, \eps) \to \R^{3 \times 3}$ with 
\begin{equation}\label{eqn:Jacobian_lower_bound}
\sigma_{\min}(\bJ(\bp_0)) > 0, 
\end{equation}
and
\[
\lim_{\kappa \to 0} \sup_{\bp \in \ball(\bp_0, \eps)} \Big\| \nabla_\bp \bF(\bp, \kappa) - \bJ(\bp) \Big\|_{\op} = 0. 
\]
\end{lemma}
\begin{lemma}\label{lem:F_kappa_derivative_solution_kappa_0}
Let Assumption \ref{assumption:activation_new} hold. Let $\bF$ be as defined in Eq. (\ref{eqn:def_bbF}). Then for any $\eps$ such that $\ball(\bp_0, 2 \eps) \subseteq \overline \Omega = \{ (\baralpha, \barsigma, \barlambda): \baralpha > - 1/\kappa, \barsigma > 0, \barlambda>0 \}$, there exists two continuous vector functions $\bF_0,\bg: \ball(\bp_0, \eps) \to \R^3$ such that
\[
\begin{aligned}
\lim_{\kappa \to 0} \sup_{\bp \in \ball(\bp_0, \eps)} \Big\| \bF(\bp, \kappa) - \bF_0(\bp) \Big\|_2 =&~ 0, \\
\lim_{\kappa \to 0} \sup_{\bp \in \ball(\bp_0, \eps)} \Big\| \partial_\kappa \bF(\bp, \kappa) - \bg(\bp) \Big\|_2 =&~ 0. 
\end{aligned}
\]
Moreover, we have 
\[
\lim_{\kappa \to 0+} \bF(\bp_0, \kappa) =  \bF_0(\bp_0) = 0.
\] 
\end{lemma}
By Lemma \ref{lem:Jacobian} and \ref{lem:F_kappa_derivative_solution_kappa_0}, we can continuously extend the function $\bF$ to the region $\ball(\bp_0, \eps) \times [0, \kappa_0)$ for some small $\kappa_0$, such that $\bF(\bp, \kappa)$ is continuously differentiable in the same region. Moreover, by Lemma \ref{lem:F_kappa_derivative_solution_kappa_0}, we have $\bF(\bp_0, 0) = \lim_{\kappa \to 0} \bF(\bp_0, \kappa) = 0$. Finally, by Lemma \ref{lem:Jacobian}, we have $\sigma_{\min}(\nabla_\bp F(\bp_0, 0)) > 0$. 

Therefore, the conditions in the Implicit Function Theorem (Lemma~\ref{lem:implicit_function}) are satisfied, from which we can conclude that there exists a continuously differentiable path $\{ \bp(\kappa); \kappa \in [0, \kappa_0) \} \subset \ball(\bp_0, \eps)$, such that $\bF(\bp(\kappa), \kappa) = 0$ for any $\kappa \in [0, \kappa_0)$. By Theorem \ref{theorem:limit} , for any $\kappa \in [0, \kappa_0)$, the equation $\bF(\bp, \kappa) = 0$ has at most one solution in $\Omega$, which should be $\bp(\kappa)$. By the differentiability of the path $\bp(\kappa)$ at $\kappa = 0$ (as a conclusion of Lemma~\ref{lem:implicit_function}), we get $\bp(\kappa)\to \bp_0$ as $\kappa\to 0$. This proves Lemma \ref{lem:small_kappa_asymptotics}. 
\qed

\subsubsection{Proof of Lemma \ref{lem:Jacobian}}
\label{appendix:proof-jacobian}
Throughout this proof, we let $Q_2=-(1 + \kappa \baralpha) Q_1 + \sqrt{\kappa} \barsigma Z$, so that $Z\sim\normal(0,1)$ and is independent of $Q_1$. We define $P = \prox_{\kappa \barlambda \rho}(Q_2)$. Then by Lemma \ref{lem:proximal}, we can define 
\[
\begin{aligned}
P_z =&~ \frac{\de}{\de z} \prox_{\kappa \barlambda \rho}(z) \vert_{z = Q_2} = 1 / (1 + \kappa \barlambda \rho''(P)), \\
P_\lambda =&~ \kappa^{-1} \frac{\de}{\de \barlambda} \prox_{\kappa \barlambda \rho}(z)\vert_{z = Q_2} = - \rho'(P) / (1 + \kappa \barlambda \rho''(P)). 
\end{aligned}
\]
The derivatives of $\bF$ gives 
\begin{align*}
\partial_{\baralpha} F_1 =&~ 4 \barlambda^2 \E[\rho'(Q_1) \rho'(P) \rho''(P) P_z Q_1] \kappa, \\
\partial_{\barsigma} F_1 =&~2 \barsigma - 4 \barlambda^2 \E[\rho'(Q_1) \rho'(P)\rho''(P) P_z Z] \sqrt{\kappa} \\
=&~ 2 \barsigma - 4 \barlambda^2 \E\Big[\rho'(Q_1) P_z \frac{\rho''(P)^2 + \kappa \barlambda \rho''(P)^3 + \rho'(P) \rho'''(P)}{(1 + \kappa \barlambda \rho''(P))^2} \Big] \barsigma \kappa \\
=&~ 2 \barsigma - 4 \barlambda^2 \E\Big[\rho'(Q_1) P_z^3 \Big(\rho''(P)^2 + \kappa \barlambda \rho''(P)^3 + \rho'(P) \rho'''(P) \Big) \Big] \barsigma \kappa, \\
\partial_{\barlambda} F_1 =&~ - 4 \barlambda \E[\rho'(Q_1) \rho'(P)^2] -4 \barlambda \E[\rho'(Q_1) \rho'(P) \rho''(P) P_\lambda] \kappa, \\
\partial_{\baralpha} F_2 =&~ - \E[\rho'(Q_1) \rho''(P) P_z Q_1^2], \\
\partial_{\barsigma} F_2 =&~ \E[\rho'(Q_1) Q_1 \rho''(P) P_z Z] \kappa^{-1/2} = \E[\rho'(Q_1) Q_1 \rho'''(P) P_z^3] \barsigma, \\
\partial_{\barlambda} F_2 =&~ \E[\rho'(Q_1) Q_1 \rho''(P) P_\lambda ], \\
\partial_{\baralpha} F_3 =&~- 2\E[\rho'(Q_1)P_z^2 \rho'''(P) P_z Q_1] \kappa \barlambda,\\
\partial_{\barsigma} F_3 =&~ 2 \E[\rho'(Q_1)P_z^2 \kappa \barlambda \rho'''(P) P_z Z] \sqrt{\kappa} \kappa^{-1}\\
=&~ 2 \E[\rho'(Q_1) (\rho''''(P) + \kappa \barlambda (\rho''(P) \rho''''(P) - 3 \rho'''(P)^2)) P_z^5] \barsigma \barlambda\kappa, \\
\partial_{\barlambda}F_3 =&~ 2\E[\rho'(Q_1)P_z^2 (\rho''(P) + \kappa \barlambda \rho'''(P) P_\lambda)] \\
=&~  2\E[\rho'(Q_1)P_z^2 \rho''(P)] + 2 \kappa \barlambda \E[\rho'(Q_1)P_z^2 \rho'''(P) P_\lambda]. 
\end{align*}
We next show that
\begin{equation}\label{eqn:F_to_J}
\begin{aligned}
\lim_{\kappa \to 0} \sup_{\bp \in \ball(\bp_0, \eps)}\| \nabla_\bp \bF(\bp, \kappa) - \bJ(\bp) \|_{\op} = 0, 
\end{aligned}
\end{equation}
where
\begin{equation}
  \label{equation:jp-expression}
\begin{aligned}
\bJ(\bp) =&~ \begin{bmatrix}
0 & 2 \barsigma & - 4 \barlambda \E[\rho'(Q_1) \rho'( - Q_1)^2]\\
- \E[\rho'(Q_1) \rho''(Q_1) Q_1^2]  & \E[\rho'(Q_1) Q_1 \rho'''(- Q_1)] \barsigma& - \E[\rho'(Q_1) \rho'(-Q_1) Q_1 \rho''(- Q_1) ] \\
0 & 0 & 2\E[\rho'(Q_1) \rho''(- Q_1)]
\end{bmatrix} \\
=&~ \begin{bmatrix}
0 & 2 \barsigma & - 4 \barlambda \E[\rho'(Q_1) \rho'(-Q_1)^2]\\
- (1/2) \E[\rho''(Q_1) Q_1^2] & \E[\rho'(Q_1) Q_1 \rho'''(-Q_1)] \barsigma & - \E[\rho'(Q_1) \rho'(-Q_1) Q_1 \rho''(Q_1) ] \\
0 & 0 &  \E[ \rho''(Q_1)]
\end{bmatrix}, \\
\end{aligned}
\end{equation}
where the last equality is by the fact that $\rho'(z) - 1/2$ is an odd function. 

The above result follows from (uniform) continuity arguments using the fact that $\rho$ is 4-th continuously differentiable: $\sup_{z} \vert \rho^{(k)}(z)\vert < \infty$ for $k = 1,2,3,4$.  In the following, we show the result for the $(3,3)$-th entry: $\lim_{\kappa \to 0} \sup_{\bp \in \ball(\bp_0, \eps)}\vert \nabla_\barlambda F_3(\bp, \kappa) - J_{33}(\bp) \vert = 0$ as a demonstration of the proof of Eq. (\ref{eqn:F_to_J}). Note we have 
\[ 
\begin{aligned}
&~\sup_{\bp \in \ball(\bp_0, \eps)}\vert \nabla_\barlambda F_3(\bp, \kappa) - J_{33}(\bp) \vert \\
\le&~\sup_{\bp \in \ball(\bp_0, \eps)} \Big\vert 2\E[\rho'(Q_1)P_z^2 \rho''(P)] + 2 \kappa \barlambda \E[\rho'(Q_1)P_z^2 \rho'''(P) P_\lambda] - 2\E[\rho'(Q_1) \rho''(- Q_1)] \Big\vert \\
\le&~\sup_{\bp \in \ball(\bp_0, \eps)} \Big\vert 2\E[\rho'(Q_1) ( P_z^2 \rho''(P) - \rho''(- Q_1) ) ] \Big\vert + \kappa \sup_{\bp \in \ball(\bp_0, \eps)} \Big\vert 2\barlambda \E[\rho'(Q_1)P_z^2 \rho'''(P) P_\lambda] \Big\vert \\
\le&~   K \cdot \E \Big[  \sup_{\bp \in \ball(\bp_0, \eps)} \vert P_z^2 \rho''(P) - \rho''(- Q_1) \vert \Big] + K \cdot \kappa  \\
\end{aligned}
\]
for some constant $K$ that does not depend on $\kappa$ and $\bp$. Note we have $\sup_{\bp \in \ball(\bp_0, \eps)} \vert P_z^2 \rho''(P) - \rho''(- Q_1) \vert$ bounded and goes to $0$ as $\kappa \to 0+$, for any fixed $(Q_1, Q_2)$. Then by bounded convergence theorem, we have 
\[ 
\begin{aligned}
\lim_{\kappa \to 0+} \sup_{\bp \in \ball(\bp_0, \eps)}\vert \nabla_\barlambda F_3(\bp, \kappa) - J_{33}(\bp) \vert = 0. 
\end{aligned}
\]
The proof of convergence of other entries within the matrix $\grad_{\bp} \bF(\bp, \kappa)-\bJ(\bp)$ follow from the same proof strategy using fact that $\sup_z|\rho^{(i)}(z)|<\infty$ for $i=1,2,3,4$.  

Finally, we show Eq. (\ref{eqn:Jacobian_lower_bound}). By Assumption~\ref{assumption:activation_new} we have $\E[ \rho''(Q_1)] > 0$, $\E[\rho''(Q_1) Q_1^2] > 0$, and $\barsigma_0 > 0$, which yields $\sigma_{\min}(\bJ(\bp_0)) > 0$. This proves Eq. (\ref{eqn:Jacobian_lower_bound}).

\qed

\subsubsection{Proof of Lemma \ref{lem:F_kappa_derivative_solution_kappa_0}}\label{appendix:proof_lem_F_kappa_derivative}


In this proof, we follow the same notations with the proof of Lemma \ref{lem:Jacobian} as in Section \ref{appendix:proof_lem_F_kappa_derivative}. Especially, the quantities $Z$, $P$, $P_z$, and $P_\lambda$ are defined accordingly. 

For any $(\baralpha, \barsigma, \barlambda, \kappa) \in \overline \Omega \times (0, \kappa_0)$ where $\overline \Omega = \{ (\baralpha, \barsigma, \barlambda): \baralpha > -1/\kappa, \barsigma > 0, \barlambda > 0\}$, we define
\begin{equation}\label{eqn:def_f_123}
\begin{aligned}
f_1(\bp, \kappa) =&~  \E[2 \rho'(Q_1) \rho'(\prox_{\kappa \barlambda \rho}(- (1 + \kappa \baralpha) Q_1 + \sqrt{\kappa} \barsigma Z))^2], \\
f_2(\bp, \kappa) =&~ \E[\rho'(Q_1) Q_1 \rho'(\prox_{\kappa \barlambda \rho}(- (1 + \kappa \baralpha) Q_1 + \sqrt{\kappa} \barsigma Z))], \\
f_3(\bp, \kappa) =&~ \E[2 \rho'(Q_1) / [1 + \kappa \barlambda \rho''(\prox_{\kappa \barlambda\rho}(- (1 + \kappa \baralpha) Q_1 + \sqrt{\kappa} \barsigma Z))] ].  
\end{aligned}
\end{equation}
By the definition of $F_1, F_2, F_3$ as in Eq. (\ref{eqn:def_bF}), we have
\begin{equation}
\begin{aligned}
F_1(\bp, \kappa) =&~ \barsigma^2 - \barlambda^2 f_1(\bp, \kappa), \\
F_2(\bp, \kappa) =&~ \kappa^{-1} f_2(\bp, \kappa), \\
F_3(\bp, \kappa) =&~ \kappa^{-1} \{1 - \kappa - f_3(\bp, \kappa) \}.  
\end{aligned}
\end{equation}
Then, Lemma \ref{lem:F_kappa_derivative_solution_kappa_0} holds as long as we show that there exists continuous functions $\bF_0(\bp) = (F_{0, 1}(\bp), F_{0, 2}(\bp), F_{0, 3}(\bp))$ and $\bg(\bp) = (g_1(\bp), g_2(\bp), g_3(\bp))$ such that
\begin{align}
f_1(\bp, \kappa) =&~  F_{0, 1}(\bp) + o(1), \label{eqn:f1_expansion_in_lemma} \\
\partial_\kappa f_1(\bp, \kappa) =&~  - (\barsigma^2 / \barlambda^2) g_1(\bp) + o(1), \label{eqn:f1_prime_expansion_in_lemma} \\
f_2(\bp, \kappa) =&~ o(1), \\
\partial_\kappa f_2(\bp, \kappa) =&~ F_{0, 2}(\bp) + o(1), \\
\partial_\kappa^2 f_2(\bp, \kappa) =&~  g_2(\bp) + o(1), \\
f_3(\bp, \kappa) =&~ 1 + o(1), \\
\partial_\kappa f_3(\bp, \kappa) =&~ - F_{0, 3}(\bp) + o(1), \\
\partial_\kappa^2 f_3(\bp, \kappa) =&~ - g_3(\bp) + o(1), 
\end{align}
where the $o(1)$ terms convergence to $0$ uniformly over $\bp \in \ball(\bp_0, \eps)$ as $\kappa \to 0+$. Moreover, we need 
\begin{align}
F_{0, 1}(\bp_0) =&~  \barsigma_0^2 / \barlambda_0^2, \label{eqn:F01_in_lemma} \\
F_{0, 2}(\bp_0) =&~ 0, \\
F_{0, 3}(\bp_0) =&~ 1.  \label{eqn:F03_in_lemma}
\end{align}

We first prove Eq. (\ref{eqn:f1_expansion_in_lemma}), (\ref{eqn:f1_prime_expansion_in_lemma}) and (\ref{eqn:F01_in_lemma}). First, we have 
\[
\lim_{\kappa \to 0+} f_1(\bp, \kappa) =  \E[2 \rho'(Q_1) \rho'(-Q_1)^2] = \barsigma_0^2 / \barlambda_0^2, 
\]
where the last equality is by the definition in Eq. (\ref{equation:bar0}). Further, by smoothness of $\rho$ and the prox operator, and the fact that the neighborhood $\ball(\bp_0, \eps)$ is bounded, this convergence is uniform over $\bp = (\baralpha, \barsigma, \barlambda) \in \ball(\bp_0, \eps)$. This proves Eq. (\ref{eqn:f1_expansion_in_lemma}) and (\ref{eqn:F01_in_lemma}). 

Second, we have 
\[
\begin{aligned}
\partial_\kappa f_1(\bp, \kappa) =&~ 4 \E\Big[ \rho'(Q_1) \rho'(P) \rho''(P) \Big( P_\lambda \barlambda - P_z \baralpha Q_1 + P_z \barsigma Z / (2 \sqrt \kappa) \Big) \Big] \\
=&~ 4 \E\Big[ \rho'(Q_1) \rho'(P) \rho''(P) \Big( P_\lambda \barlambda - P_z \baralpha Q_1 \Big) \Big] \\
&~+ 2 \E\Big[ \rho'(Q_1) [\rho''(P)^2 + \kappa \barlambda \rho''(P)^3 + \rho'(P) \rho'''(P)] P_z^3 \Big]  \barsigma^2, 
\end{aligned}
\]
where the last inequality is by Stein's identity for $Z \sim \cN(0, 1)$. So this gives 
\[
\begin{aligned}
\lim_{\kappa \to 0+} \partial_\kappa f_1(\bp, \kappa) =&~ 4 \E\Big[ \rho'(Q_1) \rho'(-Q_1) \rho''(-Q_1) \Big( - \rho'(-Q_1) \barlambda - \baralpha Q_1 \Big) \Big] \\
&~+ 2 \E\Big[ \rho'(Q_1) [\rho''(-Q_1)^2 + \rho'(-Q_1) \rho'''(-Q_1)] \Big]  \barsigma^2. 
\end{aligned}
\]
Again, by the smoothness of $\rho$ and the prox operator, and the fact that the neighborhood $\ball(\bp_0, \eps)$ is bounded, this convergence is uniform over $\bp = (\baralpha, \barsigma, \barlambda) \in \ball(\bp_0, \eps)$. This proves Eq. (\ref{eqn:f1_prime_expansion_in_lemma}). The proof of other equations within~\eqref{eqn:f1_expansion_in_lemma} to~\eqref{eqn:F03_in_lemma} follow from similar continuity arguments. This proves Lemma~\ref{lem:F_kappa_derivative_solution_kappa_0}.
\qed

%% file: Sections-arxiv/proofs.tex
\section{Proofs of main theorems}
\label{appendix:proofs}

This section provides the proofs of our main theorems, building on the developments in Section~\ref{appendix:proportional}. We first prove Theorem~\ref{theorem:general} in Section~\ref{appendix:proof-general} and prove Theorem~\ref{theorem:logistic} in Section~\ref{appendix:proof-logistic} as a direct corollary of Theorem~\ref{theorem:general}. We prove Corollary~\ref{corollary:logistic-ce} in Section~\ref{appendix:proof-logistic-ce} and Corollary~\ref{corollary:underconfidence} in Section~\ref{appendix:proof-underconfidence}.

\subsection{Proof of Theorem~\ref{theorem:general}}
\label{appendix:proof-general}

Recall from Corollary~\ref{corollary:concentration-deltapcal} that, under Assumption~\ref{assumption:activation_new}, in the limit of $d,n\to\infty$, $d/n\to\kappa$ where $\kappa<0$, the calibration error $\deltapcal(\hat{f})$ converges to the limit
\begin{align}
  \label{equation:limiting-cpkappa}
  C_{p,\kappa}(\sigma) \defeq p - \E_{Z\sim\normal(0,1)}\brac{ \sigma\paren{ \frac{\gamma}{\normbar}\cosbar \cdot \sigma^{-1}(p) + \sqrt{1-\cosbar^2}\gamma\cdot Z }},
\end{align}
where $\gamma=\norm{\wb_\star}$. This is the first part of the claim in Theorem~\ref{theorem:general}.

We now analyze the value $C_{p,\kappa}(\sigma)$ by plugging in the formulas for $\normbar,\cosbar$ from Theorem~\ref{theorem:limit}:
\begin{align*}
  \normbar = \sqrt{\alpha_\star^2\gamma^2 + \kappa\sigma_\star^2}~~~{\rm and}~~~
  \cosbar = \frac{1}{ \sqrt{1 + \kappa\sigma_\star^2/ \alpha_\star^2\gamma^2} },
\end{align*}
which yields
\begin{align*}
  \frac{\gamma\cosbar}{\normbar} = \gamma \cdot \frac{1}{\sqrt{1+\kappa\sigma_\star^2/\alpha_\star^2\gamma^2}} \cdot \frac{1}{\sqrt{\alpha_\star^2\gamma^2 + \kappa\sigma_\star^2}} = \frac{1}{\alpha_\star + \kappa\sigma_\star^2/\alpha_\star\gamma^2},
\end{align*}
and
\begin{align*}
  \sqrt{1-\cosbar^2} \cdot \gamma = \sqrt{1 - \frac{1}{1 + \kappa\sigma_\star^2 / \alpha_\star^2\gamma^2}} \cdot \gamma = \sqrt{\frac{\kappa\sigma_\star^2}{\alpha_\star^2 + \kappa\sigma_\star^2/\gamma^2}}.
\end{align*}
Now, Lemma~\ref{lem:small_kappa_asymptotics} states that the following first order approximation holds for small $\kappa$:
\begin{align*}
  & \alpha_\star = 1 + \baralpha_0\kappa + o(\kappa), \\
  & \sigma_\star^2 = \barsigma_0^2 + o(1).
\end{align*}
where $o(1)$ denotes an error term that converges to 0 as $\kappa\to 0$ (and similarly, $o(\kappa)/\kappa\to 0$). Plugging these into the preceding displays, we get
\begin{equation}
  \label{equation:first-order-1}
\begin{aligned}
  & \quad \frac{\gamma\cosbar}{\normbar} = \frac{1}{1 + \baralpha_0\kappa + o(\kappa) + \kappa(\barsigma_0^2 + o(1))/(1+o(1))\gamma^2} \\
  & = \frac{1}{1 + (\baralpha_0+\barsigma_0^2/\gamma^2)\kappa + o(\kappa)} \\
  & = \underbrace{1 - (\baralpha_0+\barsigma_0^2/\gamma^2)\kappa + o(\kappa)}_{\defeq a(\kappa)},
\end{aligned}
\end{equation}
and
\begin{align}
  \label{equation:first-order-2}
  \sqrt{1-\cosbar^2}\gamma = \sqrt{\frac{\kappa\barsigma_0^2 + o(\kappa)}{1 + \baralpha_0\kappa + o(\kappa) + \kappa(\barsigma_0^2 + o(1))/\gamma^2}} = \underbrace{\sqrt{\kappa\barsigma_0^2 + o(\kappa)}}_{\defeq b(\kappa)}.
\end{align}

We now use the following Lemma for the limiting calibration error $C_{p,\kappa}(\sigma)$, whose proof is deferred to Section~\ref{appendix:proof-cpkappa-small-kappa}:
\begin{lemma}
  \label{lemma:cpkappa-small-kappa}
  Let $C_{p,\kappa}(\sigma)$ be defined as in~\eqref{equation:limiting-cpkappa}, then we have
  \begin{align*}
    & \quad \lim_{\kappa\to 0} \frac{C_{p,\kappa}(\sigma)}{\kappa} = \lim_{\kappa\to 0} \frac{p - \E_{Z\sim\normal(0,1)}\brac{ \sigma\paren{ a(\kappa) \cdot \sigma^{-1}(p) + b(\kappa)\cdot Z }}}{\kappa} \\
    & = \underbrace{\sigma'(\sigma^{-1}(p)) \cdot \sigma^{-1}(p) \cdot \paren{\baralpha_0 + \barsigma_0^2/\gamma^2} - \frac{1}{2}\sigma''(\sigma^{-1}(p)) \cdot \barsigma_0^2}_{\defeq C(p, \baralpha_0, \barsigma_0^2,\sigma)}.
  \end{align*}
\end{lemma}
By Lemma~\ref{lemma:cpkappa-small-kappa}, the limiting analysis of $C_{p,\kappa}(\sigma)$ reduces to the analysis of the constant $C(p,\baralpha_0,\barsigma_0^2,\sigma)$. First, observe that for $p\in(0.5,1)$, by Assumption~\ref{assumption:activation_new}, we have $\sigma^{-1}(p)>0$, and (by the strict monotonicity) we have $\sigma'(\sigma^{-1}(p))>0$.

\def\climcal{C(p,\baralpha_0,\barsigma_0^2,\sigma)}

\paragraph{Proof of part (a)}
Suppose $\sigma''(\sigma^{-1}(p))\le 0$, then we have
\begin{align*}
  \climcal \ge \sigma'(\sigma^{-1}(p)) \cdot \sigma^{-1}(p) \paren{ \baralpha_0 + \barsigma_0^2/\gamma^2 }.
\end{align*}
We now prove that $\baralpha_0+\barsigma_0^2/\gamma^2>0$ always holds under Assumption~\ref{assumption:activation_new}, which implies that $\climcal\ge 0$ and thus $C_{p,\kappa}(\sigma)>0$ for sufficiently small $\kappa$. Indeed, applying the expression~\eqref{equation:baralpha0} and~\eqref{equation:barsigma0} for $\baralpha_0$, $\barsigma_0$, we have

\begin{align*}
  \baralpha_0 + \barsigma_0^2/\gamma^2 = \underbrace{\frac{\E\brac{\rho'(Q_1)\rho'(-Q_1)}}{\paren{\E\brac{\rho''(Q_1)}}^2}}_{>0} \cdot \set{-\frac{\E\brac{Q_1\rho'''(Q_1)}}{2\E\brac{Q_1^2\rho''(Q_1)} } + \frac{1}{\gamma^2}}
\end{align*}
(where $Q_1\sim\normal(0,\gamma^2)$). Therefore, it suffices to show that the quantity inside the $\set{\cdot}$ is positive, which (by $\E[Q_1^2\rho''(Q_1)]>0$, as $\rho''(\cdot)=\sigma'(\cdot)>0$) is equivalent to
\begin{align*}
  \gamma^2 \E\brac{Q_1\rho'''(Q_1)} < 2\E\brac{Q_1^2\rho''(Q_1)}.
\end{align*}
Let $Z\sim\normal(0,1)$. The above is equivalent to
\begin{align}
  \label{equation:suff-condition-stein}
  \gamma \E\brac{Z\rho'''(\gamma Z)} < 2\E\brac{Z^2\rho''(\gamma Z)}.
\end{align}
Applying Stein's identity $\E[Zf(Z)]=\E[f'(Z)]$ on $f(z)=z\rho''(\gamma z)$, we get
$
  \E\brac{Z^2\rho''(\gamma Z)} = \E\brac{\rho''(\gamma Z) + \gamma Z\rho'''(\gamma Z)},
$
which implies that $\gamma\E\brac{Z\rho'''(\gamma Z)} = \E\brac{(Z^2-1)\rho''(\gamma Z)}$. Therefore,~\eqref{equation:suff-condition-stein} is further equivalent to
\begin{align*}
  \E\brac{(Z^2-1)\rho''(\gamma Z)} < \E\brac{2Z^2\rho''(\gamma Z)},
\end{align*}
which holds as we assumed $\rho''(\cdot)=\sigma'(\cdot)>0$.

\paragraph{Proof of part (b)}
It is straightforward to see taht the following is a group a sufficient conditions for $\climcal>0$:
\begin{align*}
  \baralpha_0 \le 0~~~{\rm and}~~~\frac{1}{2}\sigma''(\sigma^{-1}(p)) > \sigma'(\sigma^{-1}(p)) \cdot \sigma^{-1}(p) /\gamma^2.
\end{align*}
Further, recall the expression~\eqref{equation:baralpha0} for $\baralpha_0$ (where $Q_1\sim\normal(0,\gamma^2)$:
\begin{align*}
  \baralpha_0 = -\frac{\E\brac{Q_1\rho'''(Q_1)} \cdot \E\brac{\rho'(Q_1)\rho'(-Q_1)}}{ 2\E\brac{Q_1^2\rho''(Q_1)} \cdot \E\brac{\rho''(Q_1)}^2}.
\end{align*}
Because $\rho'=\sigma>0$, $\rho''=\sigma'>0$, all expectations in the above expression are positive except for $\E[Q_1\rho'''(Q_1)]$. Therefore, $\baralpha_0\le 0$ is equivalent to $\E[Q_1\rho'''(Q_1)]\ge 0$. This proves $\E[Q_1\rho'''(Q_1)]\ge 0$ and $\frac{1}{2}\sigma''(\sigma^{-1}(p)) > \sigma'(\sigma^{-1}(p)) \cdot \sigma^{-1}(p) /\gamma^2$ is a sufficient condition for $\climcal<0$, which by Lemma~\ref{lemma:cpkappa-small-kappa} implies that $C_{p,\kappa}(\sigma)<0$ for all small $\kappa$, yielding part (b).
\qed

\subsubsection{Proof of Lemma~\ref{lemma:cpkappa-small-kappa}}
\label{appendix:proof-cpkappa-small-kappa}
Let $a(\kappa)=1 - (\baralpha_0+\barsigma_0^2/\gamma^2)\kappa + e(\kappa)$ and $b(\kappa)=\sqrt{\kappa\barsigma_0^2+f(\kappa)}$, where $e(\kappa), f(\kappa)=o(\kappa)$, i.e. they are bounded and satisfy $e(\kappa)/\kappa\to 0$ and $f(\kappa)/\kappa\to 0$ for small enough $\kappa$.

By a second-order Taylor expansion of $\sigma$ at $\sigma^{-1}(p)$ (which exists as Assumption~\ref{assumption:activation_new} assumed $\rho$ is four-times continuously differentiable, i.e. $\sigma$ is three-times continuously differentiable), we have
\begin{align*}
  & \quad \frac{C_{p,\kappa}(\sigma)}{\kappa} = \frac{p - \E_{Z\sim\normal(0,1)}\brac{ \sigma\paren{ a(\kappa) \cdot \sigma^{-1}(p) + b(\kappa)\cdot Z }}}{\kappa} \\
  & = \frac{p - \E_{Z\sim\normal(0,1)}\brac{ \sigma\paren{ \sigma^{-1}(p) - (\baralpha_0+\barsigma_0^2/\gamma^2)\sigma^{-1}(p)\cdot\kappa + e(\kappa)\sigma^{-1}(p) + \sqrt{\kappa\barsigma_0^2+f(\kappa)}\cdot Z }}}{\kappa} \\
  & = {\rm IV}_{\kappa} + {\rm I}_{\kappa} + {\rm II}_{\kappa} + {\rm III}_{\kappa},
\end{align*}
where terms I, II, III, IV are
\begin{align*}
  & {\rm IV}_{\kappa} = \frac{p - \sigma(\sigma^{-1}(p))}{\kappa} = 0, \\
  & {\rm I}_{\kappa} = -\frac{\sigma'(\sigma^{-1}(p)) \cdot \E\brac{ - (\baralpha_0+\barsigma_0^2/\gamma^2)\sigma^{-1}(p)\cdot\kappa + e(\kappa)\sigma^{-1}(p) + \sqrt{\kappa\barsigma_0^2+f(\kappa)}\cdot Z }}{\kappa} \\
  & \qquad = \sigma'(\sigma^{-1}(p)) \sigma^{-1}(p)\cdot (\baralpha_0+\barsigma_0^2/\gamma^2) + \sigma'(\sigma^{-1}(p)) \sigma^{-1}(p)\cdot e(\kappa)/\kappa, \\
  & {\rm II}_{\kappa} = -\frac{1}{2}\sigma''(\sigma^{-1}(p)) \cdot \frac{\E\brac{ \paren{ - (\baralpha_0+\barsigma_0^2/\gamma^2)\sigma^{-1}(p)\cdot\kappa + e(\kappa)\sigma^{-1}(p) + \sqrt{\kappa\barsigma_0^2+f(\kappa)}\cdot Z }^2 }}{\kappa}, \\
  & \abs{{\rm III}_{\kappa}} \le \E\brac{ \frac{1}{6}\abs{\sigma'''(\xi)} \cdot \abs{ - (\baralpha_0+\barsigma_0^2/\gamma^2)\sigma^{-1}(p)\cdot\kappa + e(\kappa)\sigma^{-1}(p) + \sqrt{\kappa\barsigma_0^2+f(\kappa)}\cdot Z }^3 } \bigg/ \kappa.
\end{align*}
By these expressions, at the limit $\kappa\to 0$, we have
\begin{align*}
  & \abs{{\rm III}_{\kappa}} \le \E\brac{ C_1\abs{C_2\kappa + C_3\sqrt{\kappa}Z}^3 } / \kappa \le C_4\kappa^{3/2}/\kappa = O(\sqrt{\kappa}) \to 0, \\
  & {\rm II}_{\kappa} = -\frac{1}{2}\sigma''(\sigma^{-1}(p)) \cdot \E\brac{ \paren{- (\baralpha_0+\barsigma_0^2/\gamma^2)\sigma^{-1}(p)\cdot\kappa + e(\kappa)\sigma^{-1}(p)}^2 + \paren{\kappa\barsigma_0^2 + f(\kappa)}\cdot Z^2 } \bigg/ \kappa \\
  & \qquad -\frac{1}{2}\sigma''(\sigma^{-1}(p)) \cdot \brac{ O(\kappa^2)/\kappa + (\barsigma_0^2 + f(\kappa)/\kappa) } \to -\frac{1}{2}\sigma''(\sigma^{-1}(p))\barsigma_0^2, \\
  & {\rm I}_{\kappa} \to \sigma'(\sigma^{-1}(p)) \sigma^{-1}(p)\cdot (\baralpha_0+\barsigma_0^2/\gamma^2).
\end{align*}
Therefore we have
\begin{align*}
  & \lim_{\kappa\to 0} \frac{C_{p,\kappa}(\sigma)}{\kappa} = \lim_{\kappa\to 0}{\rm I}_\kappa + \lim_{\kappa\to 0}{\rm II}_\kappa \\
  & = \sigma'(\sigma^{-1}(p)) \sigma^{-1}(p)\cdot (\baralpha_0+\barsigma_0^2/\gamma^2) -\frac{1}{2}\sigma''(\sigma^{-1}(p))\barsigma_0^2.
\end{align*}
This is the desired result.
\qed

\subsection{Proof of Theorem~\ref{theorem:logistic}}
\label{appendix:proof-logistic}
Theorem~\ref{theorem:logistic} is a special case of Theorem~\ref{theorem:general}(a). Indeed, it is straightforward to check that the logistic activation $\sigma(t)=\frac{1}{1+e^{-t}}$ along with the logistic loss $\rho(t)=\log(1+e^t)$ satisfies Assumption~\ref{assumption:activation_new}. Further, for any $t>0$, we have
\begin{align*}
  \sigma''(t) = -\frac{e^t(e^t-1)}{(1+e^t)^3} < 0.
\end{align*}
Therefore the sufficient conditions in Theorem~\ref{theorem:general}(a) is satisfied, from which we conclude that $\lim_{\kappa\to 0}C_{p,\kappa}/\kappa =C_p > 0$ for all $p\in(0.5, 1)$.
\qed

\subsection{Proof of Corollary~\ref{corollary:logistic-ce}}
\label{appendix:proof-logistic-ce}
Observe that $\hat{f}(\xb) = \sigma(\hat{\wb}^\top\xb)\eqnd \sigma(\norm{\hat{\wb}}x_1)$ where $x_1\sim\normal(0,1)$. By definition of the CE and the closed-form expression for $\deltapcal(\hat{f})$ in Lemma~\ref{lemma:calibration-error}, we have
\begin{align*}
  & \quad \CE(\hat{f}) = \E_{\xb}\brac{ \abs{\deltapcal(\hat{f})\Big|_{p=\hat{f}(\xb)}} }
    = \int_{-\infty}^\infty \abs{\deltapcal(\hat{f})\Big|_{p=\sigma(\norm{\hat{\wb}}x_1)}} \varphi(x_1)dx_1 \\
  & = \int_{-\infty}^\infty \abs{ \sigma(\norm{\hat{\wb}}x_1) - \E_{Z\sim \normal(0,1)}\brac{ \sigma\paren{ \frac{\norm{\wb_\star}}{\norm{\hat{\wb}}}\cos\hat{\theta}\cdot \sigma^{-1}(\sigma(\norm{\hat{\wb}}x_1)) + \sin\hat{\theta}\norm{\wb_\star}Z} }} \varphi(x_1) dx_1 \\
  & = \int_{-\infty}^\infty \bigg| \underbrace{ \sigma(\norm{\hat{\wb}}x_1) - \E_{Z\sim \normal(0,1)}\brac{ \sigma\paren{ \norm{\wb_\star}\cos\hat{\theta}\cdot x_1 + \sin\hat{\theta}\norm{\wb_\star}Z} }}_{\defeq g(\norm{\hat{\wb}}, \cos\hat{\theta}, x_1)} \bigg| \varphi(x_1) dx_1 \\
  & = \int_{-\infty}^\infty \abs{g(\norm{\hat{\wb}}, \cos\hat{\theta}, x_1)} \varphi(x_1) dx_1,
\end{align*}
where $\varphi(t)=\exp(-t^2/2)/\sqrt{2\pi}$ is the $\normal(0,1)$ density.

We next show that
\begin{align*}
  g(R, c, x_1) = \sigma(Rx_1) -  \E_{Z\sim \normal(0,1)}\brac{ \sigma\paren{ \gamma c\cdot x_1 + \sqrt{1-c^2}\gamma \cdot Z} }
\end{align*}
is locally Lipschitz with respect to $(R, c)$ around any $(R_0, c_0)$ such that $c_0 < 1$. ($\gamma=\norm{\wb_\star}$ for notational simplicity.) We have $|g_R'| = |\sigma'(Rx_1) \cdot x_1| \le C_1|x_1|$, and
\begin{align*}
  \abs{g_C'} = \abs{ \E_Z\brac{ \sigma'(\gamma c x_1 + \sqrt{1-c^2}\gamma Z )\cdot (\gamma x_1  -\frac{c}{\sqrt{1-c^2}}\gamma Z) }} \le C_2\paren{\abs{x_1} + \frac{1}{\sqrt{1-c_0^2}}}.
\end{align*}
Therefore we indeed have, for $(R,c)$ in a neighborhood of $(R_0, c_0)$,
\begin{align*}
  \abs{g(R_1, c_1, x_1) - g(R_2, c_2, x_1)} \le C_3\paren{\abs{x_1} + \frac{1}{\sqrt{1-c_0^2}}} \cdot \paren{ |R_1 - R_2| + |c_1 - c_2| }.
\end{align*}
Above, $C_3>0$ is an absolute constant.

Now, Theorem~\ref{theorem:limit} shows that as $d,n\to\infty$, $d/n\to\kappa$, with probability one we have $(\norm{\hat{\wb}}, \cos\hat{\theta})\to (\normbar, \cosbar)$ where $\cosbar < 1$. Therefore, we have
\begin{align*}
  & \quad \abs{ \CE(\hat{f}) - \int_{-\infty}^\infty \abs{g(\normbar, \cosbar, x_1)} \varphi(x_1) dx_1 } \\
  & \le \int_{-\infty}^\infty C_3\paren{\abs{x_1} + \frac{1}{\sqrt{1-\cosbar^2}}} \cdot \paren{ |R - R_\star| + |c - c_\star| } \varphi(x_1) dx_1 \to 0.
\end{align*}
This shows that, with probability one,
\begin{align*}
  \CE(\hat{f}) \to \int_{-\infty}^\infty \abs{g(\normbar, \cosbar, x_1)} \varphi(x_1) dx_1 = \int_{-\infty}^\infty \abs{C_{\sigma(|\normbar x_1|),\kappa}} \varphi(x_1) \cdot dx_1 \eqdef C_{\kappa} > 0.
\end{align*}
where the last step used the limiting calibration error~\eqref{equation:limiting-deltapcal}, and the fact that for $p<0.5$ we have $|\deltapcal(\hat{f})|=|\Delta_{1-p}^{\sf cal}(\hat{f})|$ by symmetry of the activation function. This shows the first part of the corollary.

Finally, Theorem~\ref{theorem:logistic} asserts that
$C_{p,\kappa} = C_p\kappa + o(\kappa)$ for sufficiently small $\kappa$. Further, this $o(\kappa)$ by the proof of Lemma~\ref{lemma:cpkappa-small-kappa} is uniform over $p\in(0.5, 1)$, provided that $\sup_z \{|\sigma'(z)z|, |\sigma''(z)z^2|, |\sigma'''(z) \cdot \E_{G\sim \normal(0,1)}[|z + aG|^3]|\}<\infty$. For the logistic activation $\sigma(z)=1/(1+e^{-z})$ these are all satisfied due to the exponential decay of $\sigma',\sigma'',\sigma'''$. This means we can plug in $C_{p,\kappa}=C_p\kappa$ into the above and obtain
\begin{align*}
  C_\kappa = \underbrace{\int_{-\infty}^\infty C_{\sigma(|\normbar x_1|)} \varphi(x_1)dx_1}_{\defeq C} \cdot \kappa + o(\kappa),
\end{align*}
where $C>0$. This shows the second part of the corollary.
\qed

\subsection{Proof of Corollary~\ref{corollary:underconfidence}}
\label{appendix:proof-underconfidence}
We first check that $\sigma_{\uc}$ defined in~\eqref{equation:cos-activation} satisfy the sufficient conditions of Theorem~\ref{theorem:general}(b) at $\norm{\wb_\star}=1$. Recall
\begin{equation}
  \sigma_{\uc}(z) = \left\{
  \begin{aligned}
    & 0, & z < -2\pi, \\
    & \frac{1}{2} + \frac{1}{4\pi}(z - \sin z), & |z| \le 2\pi, \\
    & 1, & z > 2\pi.
  \end{aligned}
  \right.
\end{equation}
First, through numerical calculations, we get
\begin{align*}
  \E_{Q_1\sim \normal(0,1)} \brac{ Q_1\sigma''(Q_1)} = \E_{Q_1\sim\normal(0,1)} \brac{Q_1 \cdot \frac{1}{4\pi}\sin(Q_1)} = 0.0483 > 0.
\end{align*}
This verifies condition~\eqref{equation:suff-uc-1}. Second, for $z=\sigma^{-1}(p)$ and $|z|\le 2\pi$ we have
\begin{align*}
  \sigma_{\uc}''(z) - 2\sigma_{\uc}'(z)\cdot z = \frac{1}{4\pi}\paren{ \sin z + 2z\cos z - 2z }.
\end{align*}
Numerical calculation shows that the above is strictly positive for (at least) $z\in(0, 0.96]$. This shows that the under-confidence provably happens at $p\in(0.5, \sigma(0.96)]=(0.5, 0.5112]$ in the limit of $\kappa\to 0$. This verified condition~\eqref{equation:suff-uc-2} for this range of $p$ (and sufficiently small $\kappa$). We remark in passing that the upper range $p\approx 0.5112$ of under-confidence agrees well with the simulations at $d/n=0.01$ in Figure~\ref{figure:simulation}(c).

Finally, note that $\sigma_{\uc}$ satisfies all symmetry and low-order smoothness assumptions in Assumption~\ref{assumption:activation_new}, but does not satisfy the high-order smoothness ($\sigma''=\rho'''$ is still continuous, but $\sigma'''=\rho''''$ does not exist) as well as the strict positivity $\sigma'(z)>0$. However, since $\sigma_{\uc}$ satisfy the above two sufficient conditions with a positive margin, we can slightly perturb and smoothify it to some $\widetilde{\sigma}_{\uc}$ which does satisfy higher-order smoothness and strict positivity, and preserves the above two sufficient conditions. This proves Corollary~\ref{corollary:underconfidence}.
\qed

%% file: Sections-arxiv/appendix-experiments.tex
\section{Additional experimental details}
\label{appendix:experiments}

\subsection{Simulations}
\label{appendix:simulations-details}

We choose $d=100$ and $d/n\in\set{0.01, 0.05, 0.1, 0.25}$ which corresponds to $n\in\set{10000, 2000, 1000, 400}$. For all settings we generalize $5$ problem instances. We solve the convex ERM (with either $\sigma_{\logistic}$ or $\sigma_{\uc}$) with gradient descent. We run gradient descent with step size $0.01$ on the full loss $\hat{R}_n(\wb)$, until the gradient norm $\norm{\grad \hat{R}_n(\wb)}<10^{-5}$ (as $\hat{R}_n$ is convex, gradient norm is a proper measure of global optimality). Each problem instance yields a $\hat{\wb}$ for which we can compute $\norm{\hat{\wb}}$ and $\cos\hat{\theta}=\hat{\wb}^\top\wb_\star/\norm{\hat{\wb}}\norm{\wb_\star}$, and then compute the calibration curve from the closed-form expression~\eqref{equation:deltapcal-expression}, as shown in Figure~\ref{figure:simulation}.

\subsection{CIFAR10 experiment}
\label{appendix:cifar10-details}
We train a multi-class logistic (softmax) classifier $\hat{\Wb}\in\R^{3072 \times 5}$ for the ($5$-class subset of) CIFAR10 data. Note that the classifier does not involve an intercept (which does not restrict the capacity of the model as the data is class-balanced). We train on the training set (with $n=25000$ examples) with the momentum optimizer with learning rate $\set{10^{-3}, 10^{-4}, 10^{-5}}$ for $30$ epochs each (totally $90$ epochs), and momentum $0.9$. We use minibatch gradients with batch-size  $128$. The training and test images are pre-processed by subtracting the mean and normalizing by the standard deviation, and {\bf without} any further data augmentation.

The trained $\hat{\Wb}$ is then used as-is to generate (random) pseudo-labels $y_i^{\sf pseudo}$ as described in Section~\ref{section:experiments-real}, for both the training and test sets. We then use the same training setup on the pseudo training set $(\xb_i, y_i^{\sf pseudo})$ to learn a logistic classifier $\wt{\Wb}\in\R^{3072\times 5}$, and evaluate its accuracy and confidence on the test (pseudo-labeled) dataset.

\begin{figure}[th]
  \centering
  \includegraphics[width=0.45\textwidth]{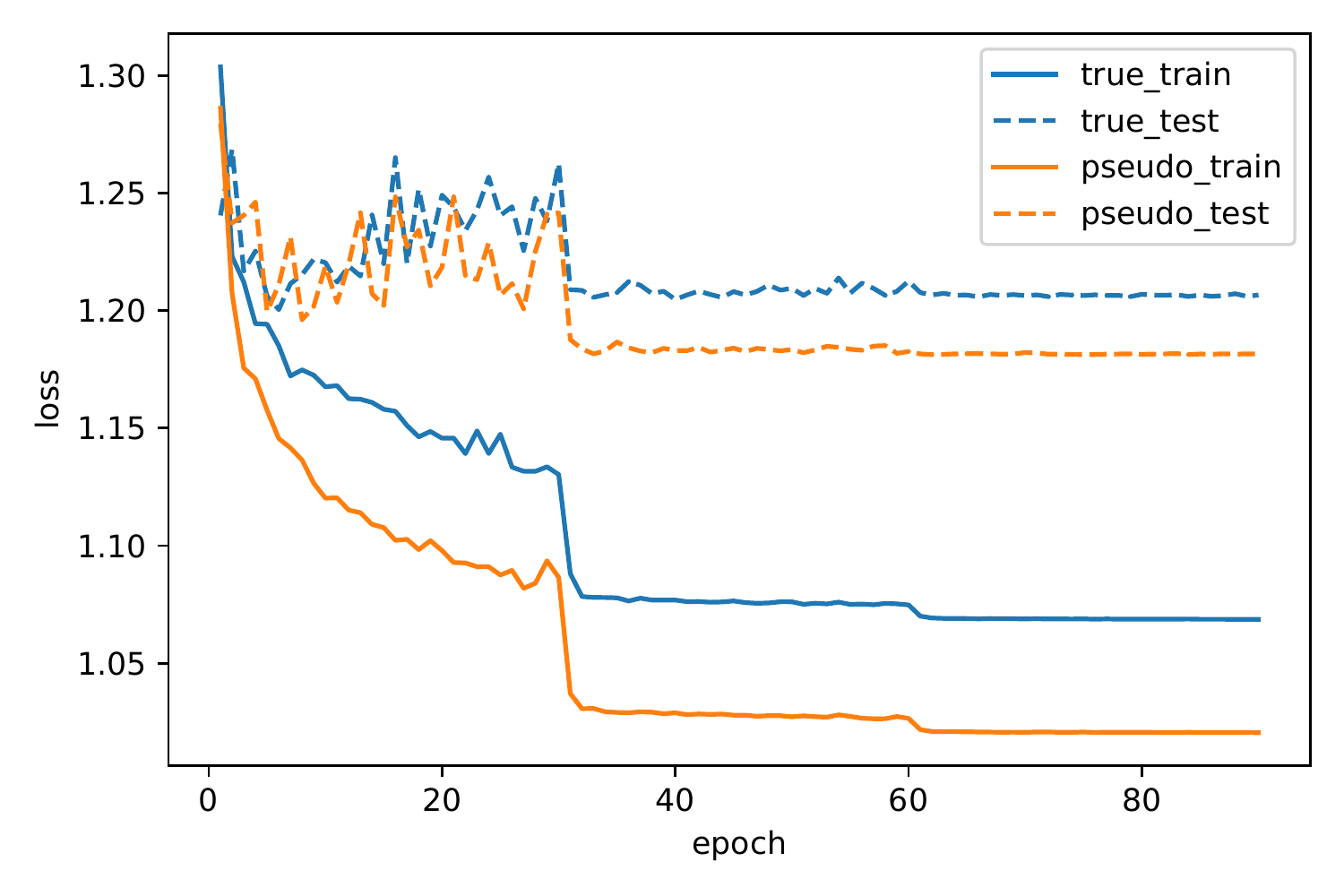}
  \quad
  \includegraphics[width=0.45\textwidth]{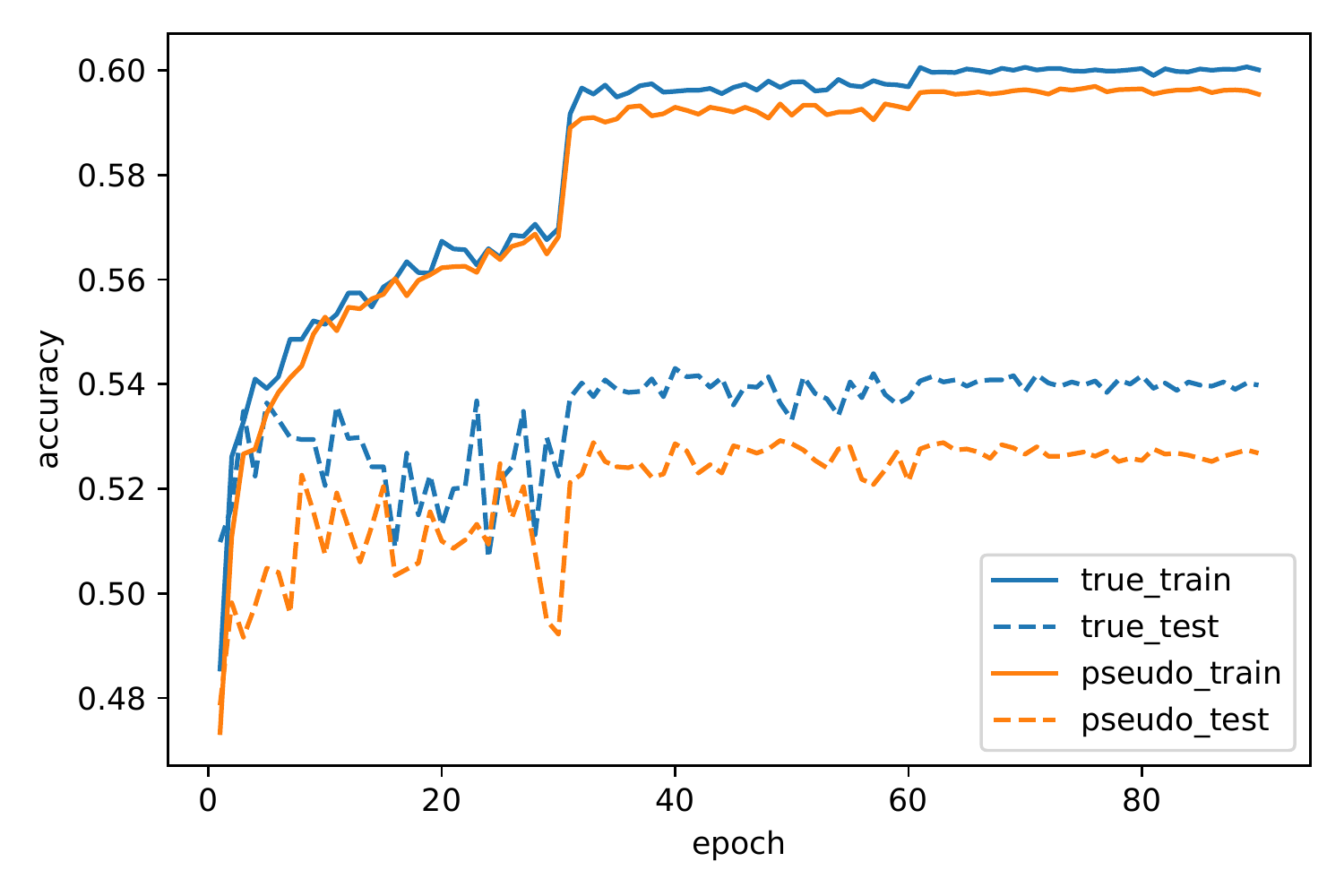}
  \caption{Training and test losses (left) and accuracies (right) for the CIFAR10 logistic regression experiment in Section~\ref{section:experiments-real}.}
  \label{figure:cifar-losses}
\end{figure}

The training and test losses for both the true labels and the pseudo labels are plotted in Figure~\ref{figure:cifar-losses}. As indicated by the final training loss, the problem is under-parametrized (training loss is well above $0$) for both tasks. Further, observe that both losses have stabilized at the end of training.

For evaluating the confidence and accuracies on both the true labels and pseudo labels, we partition $[0.2, 10]$ to 10 equally spaced bins $B_1,\dots,B_{10}$, and compute its average confidence and accuracy within $B_j$ on the test set as
\begin{align*}
  & \overline{\rm conf}_j = \frac{1}{|B_j|} \sum_{i:{\rm conf}_i\in B_j} {\rm conf}_i. \\
  & \overline{\rm acc}_j = \frac{1}{|B_j|} \sum_{i: {\rm conf}_i \in B_j} \indic{\argmax_k \brac{\wt{\Wb}^\top \xb_i}_k = y_i^{\sf pseudo} },
\end{align*}
where
\begin{align*}
  {\rm conf}_i = \max_k \frac{\exp\paren{\brac{\wt{\Wb}^\top \xb_i}_k}}{\sum_{k'=1}^5 \exp\paren{\brac{\wt{\Wb}^\top \xb_i}_{k'}}} \in [0.2, 1]
\end{align*}
is the confidence on the $i$-th test example for $i\in\set{1,\dots,5000}$. The $\overline{\rm conf}_j$ and $\overline{\rm acc}_j$ are plotted in the calibration diagrams in Figure~\ref{figure:cifar10}.

\subsection{An ablation}
\begin{figure*}[h]
  \centering
  \begin{minipage}{0.49\textwidth}
    \centering
    \subcaption{True labels}\label{figure:cifar-true}
    \vspace{-.7em}
    \includegraphics[width=0.49\textwidth]{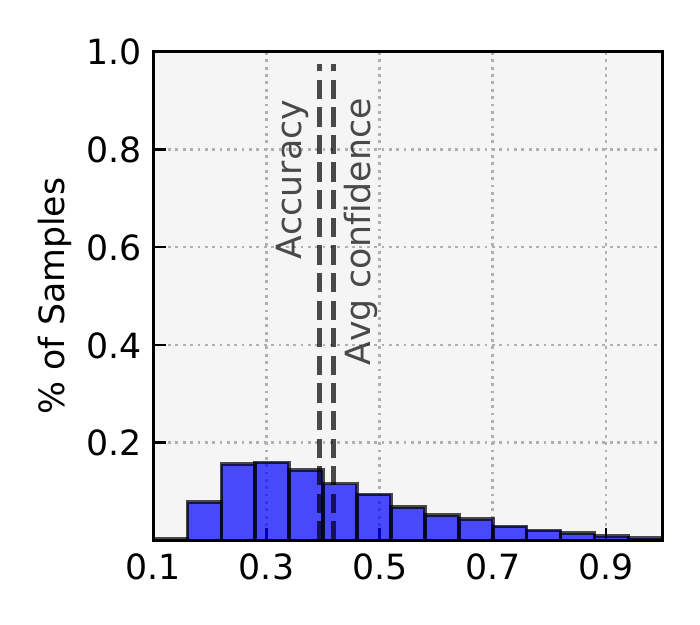}
    \includegraphics[width=0.49\textwidth]{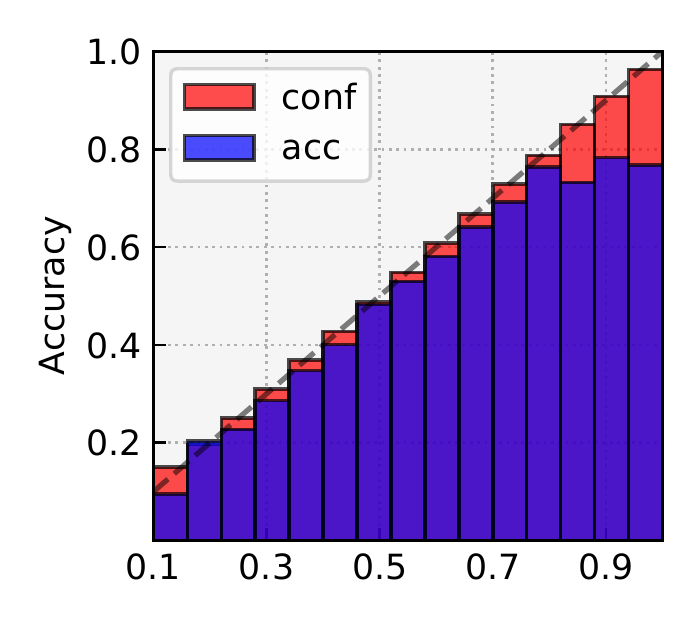}
  \end{minipage}
  \begin{minipage}{0.49\textwidth}
    \centering
    \subcaption{Pseudo labels (realizable)}\label{figure:cifar-pseudo}
    \vspace{-.7em}
    \includegraphics[width=0.49\textwidth]{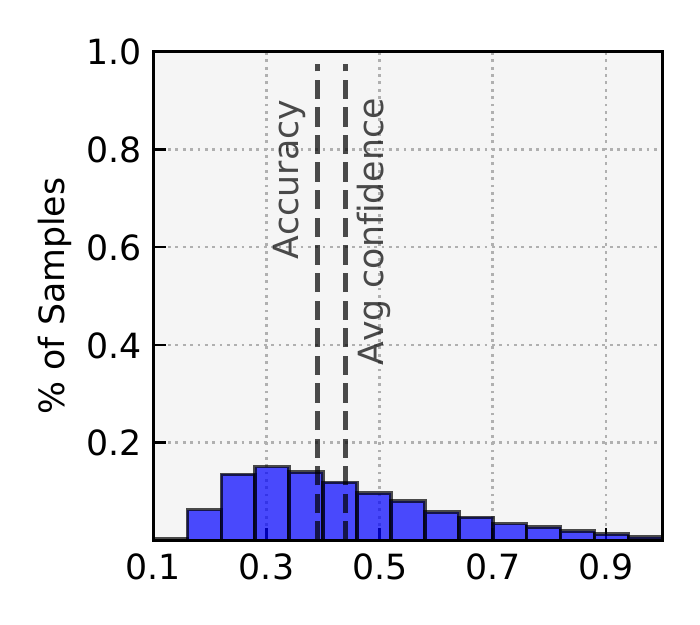}
    \includegraphics[width=0.49\textwidth]{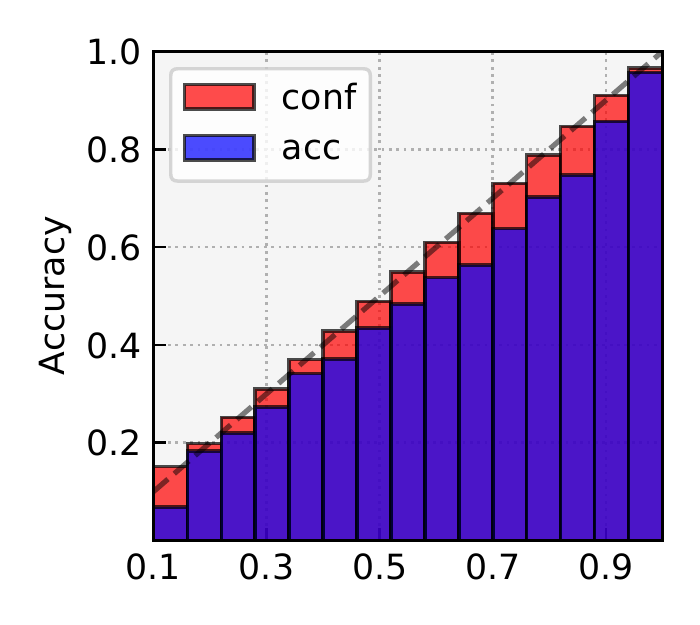}
  \end{minipage}
  \vspace{-1em}
  \caption{\small Calibration of multi-class logistic regression on (all 10 classes of) CIFAR10. Confidences are partitioned into $B=15$ bins. The $x$-axes denote the confidences (predicted top probabilities) of the models.}
  \label{figure:cifar10-ablation}
\end{figure*}

Figure~\ref{figure:cifar10-ablation} reports the results of re-running the CIFAR experiment in Section~\ref{section:experiments-real}, but with the full 10 classes of data (instead of 5), and 15 confidence bins (instead of 10). The test accuracy of logistic regression is now $40\%$. We find that logistic regression is over-confident on both true labels and pseudo-labels, similar as in Section~\ref{section:experiments-real}.

\subsection{Details of Figure~\ref{figure:fig1}}
The deep neural net presented in Figure~\ref{figure:fig1} is a WideResNet-50-2~\citep{zagoruyko2016wide} trained on the ImageNet image classification dataset. We train for 100 epochs with the momentum optimizer with Nesterov momentum 0.9, initial learning rate $0.1$, a learning rate decay factor of $10\times$ at the $\set{30, 60, 90}$ epochs, and weight decay $5\times 10^{-4}$. We use a batchsize of $256$ distributed to $8$ GPUs. The final test accuracy of this model is $76.27\%$. The realizable logistic regression uses simulated data with $n=2000,d=100$, and the same training protocol as described in Section~\ref{appendix:simulations-details}.

%% file: calibration-arxiv.bbl
\begin{thebibliography}{48}
\providecommand{\natexlab}[1]{#1}
\providecommand{\url}[1]{\texttt{#1}}
\expandafter\ifx\csname urlstyle\endcsname\relax
  \providecommand{\doi}[1]{doi: #1}\else
  \providecommand{\doi}{doi: \begingroup \urlstyle{rm}\Url}\fi

\bibitem[Albert and Anderson(1984)]{albert1984existence}
Adelin Albert and John~A Anderson.
\newblock On the existence of maximum likelihood estimates in logistic
  regression models.
\newblock \emph{Biometrika}, 71\penalty0 (1):\penalty0 1--10, 1984.

\bibitem[Bayati and Montanari(2011)]{bayati2011dynamics}
Mohsen Bayati and Andrea Montanari.
\newblock The dynamics of message passing on dense graphs, with applications to
  compressed sensing.
\newblock \emph{IEEE Transactions on Information Theory}, 57\penalty0
  (2):\penalty0 764--785, 2011.

\bibitem[Begoli et~al.(2019)Begoli, Bhattacharya, and Kusnezov]{begoli2019need}
Edmon Begoli, Tanmoy Bhattacharya, and Dimitri Kusnezov.
\newblock The need for uncertainty quantification in machine-assisted medical
  decision making.
\newblock \emph{Nature Machine Intelligence}, 1\penalty0 (1):\penalty0 20--23,
  2019.

\bibitem[Cand{\`e}s et~al.(2020)Cand{\`e}s, Sur, et~al.]{candes2020phase}
Emmanuel~J Cand{\`e}s, Pragya Sur, et~al.
\newblock The phase transition for the existence of the maximum likelihood
  estimate in high-dimensional logistic regression.
\newblock \emph{The Annals of Statistics}, 48\penalty0 (1):\penalty0 27--42,
  2020.

\bibitem[Ding et~al.(2020)Ding, Han, Liu, and Niethammer]{ding2020local}
Zhipeng Ding, Xu~Han, Peirong Liu, and Marc Niethammer.
\newblock Local temperature scaling for probability calibration.
\newblock \emph{arXiv preprint arXiv:2008.05105}, 2020.

\bibitem[Donoho et~al.(2009)Donoho, Maleki, and Montanari]{donoho2009message}
David~L Donoho, Arian Maleki, and Andrea Montanari.
\newblock Message-passing algorithms for compressed sensing.
\newblock \emph{Proceedings of the National Academy of Sciences}, 106\penalty0
  (45):\penalty0 18914--18919, 2009.

\bibitem[Dusenberry et~al.(2020)Dusenberry, Jerfel, Wen, Ma, Snoek, Heller,
  Lakshminarayanan, and Tran]{dusenberry2020efficient}
Michael Dusenberry, Ghassen Jerfel, Yeming Wen, Yian Ma, Jasper Snoek,
  Katherine Heller, Balaji Lakshminarayanan, and Dustin Tran.
\newblock Efficient and scalable bayesian neural nets with rank-1 factors.
\newblock In \emph{International conference on machine learning}, pages
  2782--2792. PMLR, 2020.

\bibitem[El~Karoui et~al.(2013)El~Karoui, Bean, Bickel, Lim, and
  Yu]{el2013robust}
Noureddine El~Karoui, Derek Bean, Peter~J Bickel, Chinghway Lim, and Bin Yu.
\newblock On robust regression with high-dimensional predictors.
\newblock \emph{Proceedings of the National Academy of Sciences}, 110\penalty0
  (36):\penalty0 14557--14562, 2013.

\bibitem[Gal and Ghahramani(2016)]{gal2016dropout}
Yarin Gal and Zoubin Ghahramani.
\newblock Dropout as a bayesian approximation: Representing model uncertainty
  in deep learning.
\newblock In \emph{international conference on machine learning}, pages
  1050--1059. PMLR, 2016.

\bibitem[Gal et~al.(2017)Gal, Hron, and Kendall]{gal2017concrete}
Yarin Gal, Jiri Hron, and Alex Kendall.
\newblock Concrete dropout.
\newblock \emph{arXiv preprint arXiv:1705.07832}, 2017.

\bibitem[Guo et~al.(2017)Guo, Pleiss, Sun, and Weinberger]{guo2017calibration}
Chuan Guo, Geoff Pleiss, Yu~Sun, and Kilian~Q Weinberger.
\newblock On calibration of modern neural networks.
\newblock \emph{arXiv preprint arXiv:1706.04599}, 2017.

\bibitem[Gupta et~al.(2020)Gupta, Podkopaev, and Ramdas]{gupta2020distribution}
Chirag Gupta, Aleksandr Podkopaev, and Aaditya Ramdas.
\newblock Distribution-free binary classification: prediction sets, confidence
  intervals and calibration.
\newblock \emph{arXiv preprint arXiv:2006.10564}, 2020.

\bibitem[Hastie et~al.(2009)Hastie, Tibshirani, and
  Friedman]{hastie2009elements}
Trevor Hastie, Robert Tibshirani, and Jerome Friedman.
\newblock \emph{The elements of statistical learning: data mining, inference,
  and prediction}.
\newblock Springer Science \& Business Media, 2009.

\bibitem[Jung et~al.(2020)Jung, Lee, Pai, Roth, and Vohra]{jung2020moment}
Christopher Jung, Changhwa Lee, Mallesh~M Pai, Aaron Roth, and Rakesh Vohra.
\newblock Moment multicalibration for uncertainty estimation.
\newblock \emph{arXiv preprint arXiv:2008.08037}, 2020.

\bibitem[Kakade et~al.(2011)Kakade, Kalai, Kanade, and
  Shamir]{kakade2011efficient}
Sham Kakade, Adam~Tauman Kalai, Varun Kanade, and Ohad Shamir.
\newblock Efficient learning of generalized linear and single index models with
  isotonic regression.
\newblock \emph{arXiv preprint arXiv:1104.2018}, 2011.

\bibitem[Karoui(2013)]{karoui2013asymptotic}
Noureddine~El Karoui.
\newblock Asymptotic behavior of unregularized and ridge-regularized
  high-dimensional robust regression estimators: rigorous results.
\newblock \emph{arXiv preprint arXiv:1311.2445}, 2013.

\bibitem[Kull et~al.(2017)Kull, Silva~Filho, and Flach]{kull2017beta}
Meelis Kull, Telmo Silva~Filho, and Peter Flach.
\newblock Beta calibration: a well-founded and easily implemented improvement
  on logistic calibration for binary classifiers.
\newblock In \emph{Artificial Intelligence and Statistics}, pages 623--631.
  PMLR, 2017.

\bibitem[Kull et~al.(2019)Kull, Perello-Nieto, K{\"a}ngsepp, Song, Flach,
  et~al.]{kull2019beyond}
Meelis Kull, Miquel Perello-Nieto, Markus K{\"a}ngsepp, Hao Song, Peter Flach,
  et~al.
\newblock Beyond temperature scaling: Obtaining well-calibrated multiclass
  probabilities with dirichlet calibration.
\newblock \emph{arXiv preprint arXiv:1910.12656}, 2019.

\bibitem[Kumar et~al.(2019)Kumar, Liang, and Ma]{kumar2019verified}
Ananya Kumar, Percy Liang, and Tengyu Ma.
\newblock Verified uncertainty calibration.
\newblock \emph{arXiv preprint arXiv:1909.10155}, 2019.

\bibitem[Lakshminarayanan et~al.(2016)Lakshminarayanan, Pritzel, and
  Blundell]{lakshminarayanan2016simple}
Balaji Lakshminarayanan, Alexander Pritzel, and Charles Blundell.
\newblock Simple and scalable predictive uncertainty estimation using deep
  ensembles.
\newblock \emph{arXiv preprint arXiv:1612.01474}, 2016.

\bibitem[Liu et~al.(2020)Liu, Lin, Padhy, Tran, Bedrax-Weiss, and
  Lakshminarayanan]{liu2020simple}
Jeremiah~Zhe Liu, Zi~Lin, Shreyas Padhy, Dustin Tran, Tania Bedrax-Weiss, and
  Balaji Lakshminarayanan.
\newblock Simple and principled uncertainty estimation with deterministic deep
  learning via distance awareness.
\newblock \emph{arXiv preprint arXiv:2006.10108}, 2020.

\bibitem[Liu et~al.(2019)Liu, Simchowitz, and Hardt]{liu2019implicit}
Lydia~T Liu, Max Simchowitz, and Moritz Hardt.
\newblock The implicit fairness criterion of unconstrained learning.
\newblock In \emph{International Conference on Machine Learning}, pages
  4051--4060. PMLR, 2019.

\bibitem[Maddox et~al.(2019)Maddox, Izmailov, Garipov, Vetrov, and
  Wilson]{maddox2019simple}
Wesley~J Maddox, Pavel Izmailov, Timur Garipov, Dmitry~P Vetrov, and
  Andrew~Gordon Wilson.
\newblock A simple baseline for bayesian uncertainty in deep learning.
\newblock \emph{Advances in Neural Information Processing Systems},
  32:\penalty0 13153--13164, 2019.

\bibitem[Mai et~al.(2019)Mai, Liao, and Couillet]{mai2019large}
Xiaoyi Mai, Zhenyu Liao, and Romain Couillet.
\newblock A large scale analysis of logistic regression: Asymptotic performance
  and new insights.
\newblock In \emph{ICASSP 2019-2019 IEEE International Conference on Acoustics,
  Speech and Signal Processing (ICASSP)}, pages 3357--3361. IEEE, 2019.

\bibitem[Malinin et~al.(2019)Malinin, Mlodozeniec, and
  Gales]{malinin2019ensemble}
Andrey Malinin, Bruno Mlodozeniec, and Mark Gales.
\newblock Ensemble distribution distillation.
\newblock \emph{arXiv preprint arXiv:1905.00076}, 2019.

\bibitem[McCullagh(2018)]{mccullagh2018generalized}
Peter McCullagh.
\newblock \emph{Generalized linear models}.
\newblock Routledge, 2018.

\bibitem[Michelmore et~al.(2018)Michelmore, Kwiatkowska, and
  Gal]{michelmore2018evaluating}
Rhiannon Michelmore, Marta Kwiatkowska, and Yarin Gal.
\newblock Evaluating uncertainty quantification in end-to-end autonomous
  driving control.
\newblock \emph{arXiv preprint arXiv:1811.06817}, 2018.

\bibitem[Mukhoti et~al.(2020)Mukhoti, Kulharia, Sanyal, Golodetz, Torr, and
  Dokania]{mukhoti2020calibrating}
Jishnu Mukhoti, Viveka Kulharia, Amartya Sanyal, Stuart Golodetz, Philip~HS
  Torr, and Puneet~K Dokania.
\newblock Calibrating deep neural networks using focal loss.
\newblock \emph{arXiv preprint arXiv:2002.09437}, 2020.

\bibitem[Naeini et~al.(2015)Naeini, Cooper, and
  Hauskrecht]{naeini2015obtaining}
Mahdi~Pakdaman Naeini, Gregory Cooper, and Milos Hauskrecht.
\newblock Obtaining well calibrated probabilities using bayesian binning.
\newblock In \emph{Proceedings of the AAAI Conference on Artificial
  Intelligence}, volume~29, 2015.

\bibitem[Nixon et~al.(2019)Nixon, Dusenberry, Zhang, Jerfel, and
  Tran]{nixon2019measuring}
Jeremy Nixon, Michael~W Dusenberry, Linchuan Zhang, Ghassen Jerfel, and Dustin
  Tran.
\newblock Measuring calibration in deep learning.
\newblock In \emph{CVPR Workshops}, volume~2, 2019.

\bibitem[Ovadia et~al.(2019)Ovadia, Fertig, Ren, Nado, Sculley, Nowozin,
  Dillon, Lakshminarayanan, and Snoek]{ovadia2019can}
Yaniv Ovadia, Emily Fertig, Jie Ren, Zachary Nado, David Sculley, Sebastian
  Nowozin, Joshua Dillon, Balaji Lakshminarayanan, and Jasper Snoek.
\newblock Can you trust your model's uncertainty? evaluating predictive
  uncertainty under dataset shift.
\newblock In \emph{Advances in Neural Information Processing Systems}, pages
  13991--14002, 2019.

\bibitem[Platt et~al.(1999)]{platt1999probabilistic}
John Platt et~al.
\newblock Probabilistic outputs for support vector machines and comparisons to
  regularized likelihood methods.
\newblock 1999.

\bibitem[Rahimi et~al.(2020)Rahimi, Shaban, Cheng, Hartley, and
  Boots]{rahimi2020intra}
Amir Rahimi, Amirreza Shaban, Ching-An Cheng, Richard Hartley, and Byron Boots.
\newblock Intra order-preserving functions for calibration of multi-class
  neural networks.
\newblock \emph{Advances in Neural Information Processing Systems}, 33, 2020.

\bibitem[Shabat et~al.(2020)Shabat, Cohen, and Mansour]{shabat2020sample}
Eliran Shabat, Lee Cohen, and Yishay Mansour.
\newblock Sample complexity of uniform convergence for multicalibration.
\newblock \emph{arXiv preprint arXiv:2005.01757}, 2020.

\bibitem[Soudry et~al.(2018)Soudry, Hoffer, Nacson, Gunasekar, and
  Srebro]{soudry2018implicit}
Daniel Soudry, Elad Hoffer, Mor~Shpigel Nacson, Suriya Gunasekar, and Nathan
  Srebro.
\newblock The implicit bias of gradient descent on separable data.
\newblock \emph{The Journal of Machine Learning Research}, 19\penalty0
  (1):\penalty0 2822--2878, 2018.

\bibitem[Stojnic(2013)]{stojnic2013framework}
Mihailo Stojnic.
\newblock A framework to characterize performance of lasso algorithms.
\newblock \emph{arXiv preprint arXiv:1303.7291}, 2013.

\bibitem[Sur and Cand{\`e}s(2019)]{sur2019modern}
Pragya Sur and Emmanuel~J Cand{\`e}s.
\newblock A modern maximum-likelihood theory for high-dimensional logistic
  regression.
\newblock \emph{Proceedings of the National Academy of Sciences}, 116\penalty0
  (29):\penalty0 14516--14525, 2019.

\bibitem[Taheri et~al.(2020)Taheri, Pedarsani, and
  Thrampoulidis]{taheri2020fundamental}
Hossein Taheri, Ramtin Pedarsani, and Christos Thrampoulidis.
\newblock Fundamental limits of ridge-regularized empirical risk minimization
  in high dimensions.
\newblock \emph{arXiv preprint arXiv:2006.08917}, 2020.

\bibitem[Thrampoulidis et~al.(2015)Thrampoulidis, Oymak, and
  Hassibi]{thrampoulidis2015regularized}
Christos Thrampoulidis, Samet Oymak, and Babak Hassibi.
\newblock Regularized linear regression: A precise analysis of the estimation
  error.
\newblock In \emph{Conference on Learning Theory}, pages 1683--1709. PMLR,
  2015.

\bibitem[Thrampoulidis et~al.(2018)Thrampoulidis, Abbasi, and
  Hassibi]{thrampoulidis2018precise}
Christos Thrampoulidis, Ehsan Abbasi, and Babak Hassibi.
\newblock Precise error analysis of regularized $ m $-estimators in high
  dimensions.
\newblock \emph{IEEE Transactions on Information Theory}, 64\penalty0
  (8):\penalty0 5592--5628, 2018.

\bibitem[Thulasidasan et~al.(2019)Thulasidasan, Chennupati, Bilmes,
  Bhattacharya, and Michalak]{thulasidasan2019mixup}
Sunil Thulasidasan, Gopinath Chennupati, Jeff Bilmes, Tanmoy Bhattacharya, and
  Sarah Michalak.
\newblock On mixup training: Improved calibration and predictive uncertainty
  for deep neural networks.
\newblock \emph{arXiv preprint arXiv:1905.11001}, 2019.

\bibitem[Tran et~al.(2020)Tran, Veeling, Roth, Swiatkowski, Dillon, Snoek,
  Mandt, Salimans, Nowozin, and Jenatton]{tran2020hydra}
Linh Tran, Bastiaan~S Veeling, Kevin Roth, Jakub Swiatkowski, Joshua~V Dillon,
  Jasper Snoek, Stephan Mandt, Tim Salimans, Sebastian Nowozin, and Rodolphe
  Jenatton.
\newblock Hydra: Preserving ensemble diversity for model distillation.
\newblock \emph{arXiv preprint arXiv:2001.04694}, 2020.

\bibitem[Van~der Vaart(2000)]{van2000asymptotic}
Aad~W Van~der Vaart.
\newblock \emph{Asymptotic statistics}, volume~3.
\newblock Cambridge university press, 2000.

\bibitem[Wen et~al.(2020)Wen, Tran, and Ba]{wen2020batchensemble}
Yeming Wen, Dustin Tran, and Jimmy Ba.
\newblock Batchensemble: an alternative approach to efficient ensemble and
  lifelong learning.
\newblock In \emph{International Conference on Learning Representations}, 2020.
\newblock URL \url{https://openreview.net/forum?id=Sklf1yrYDr}.

\bibitem[Zadrozny and Elkan()]{zadrozny2001obtaining}
Bianca Zadrozny and Charles Elkan.
\newblock Obtaining calibrated probability estimates from decision trees and
  naive bayesian classifiers.
\newblock Citeseer.

\bibitem[Zadrozny and Elkan(2002)]{zadrozny2002transforming}
Bianca Zadrozny and Charles Elkan.
\newblock Transforming classifier scores into accurate multiclass probability
  estimates.
\newblock In \emph{Proceedings of the eighth ACM SIGKDD international
  conference on Knowledge discovery and data mining}, pages 694--699, 2002.

\bibitem[Zagoruyko and Komodakis(2016)]{zagoruyko2016wide}
Sergey Zagoruyko and Nikos Komodakis.
\newblock Wide residual networks.
\newblock \emph{arXiv preprint arXiv:1605.07146}, 2016.

\bibitem[Zhang et~al.(2020)Zhang, Kailkhura, and Han]{zhang2020mix}
Jize Zhang, Bhavya Kailkhura, and T~Yong-Jin Han.
\newblock Mix-n-match: Ensemble and compositional methods for uncertainty
  calibration in deep learning.
\newblock In \emph{International Conference on Machine Learning}, pages
  11117--11128. PMLR, 2020.

\end{thebibliography}
